\pdfoutput=1
\documentclass{article}
\usepackage[T1]{fontenc}
\usepackage{amsmath,amssymb,graphicx,booktabs,caption,subcaption,makecell,xcolor,longtable,array}
\usepackage[final]{corl_2026}

\title{Box Trajectories as Compact State Representations for World Action Models}
\author{
Qinxi Yu \quad Wenzhou Ding \quad Gloria X. Wang \\
\textbf{Albert J. Zhai \quad Shenlong Wang} \\
University of Illinois Urbana-Champaign \\
}
\newcommand{\methodname}{BoxWAM}
\newcommand{\method}{BoxWAM}

\hypersetup{
 pdftitle={Box Trajectories as Compact State Representations for World Action Models},
 pdfauthor={Qinxi Yu, Wenzhou Ding, Gloria X. Wang, Albert J. Zhai, Shenlong Wang}
}
\begin{document}
\maketitle
\begin{center}
\nopagebreak
\vspace{-1.8em}
\noindent
\includegraphics[width=0.96\textwidth]{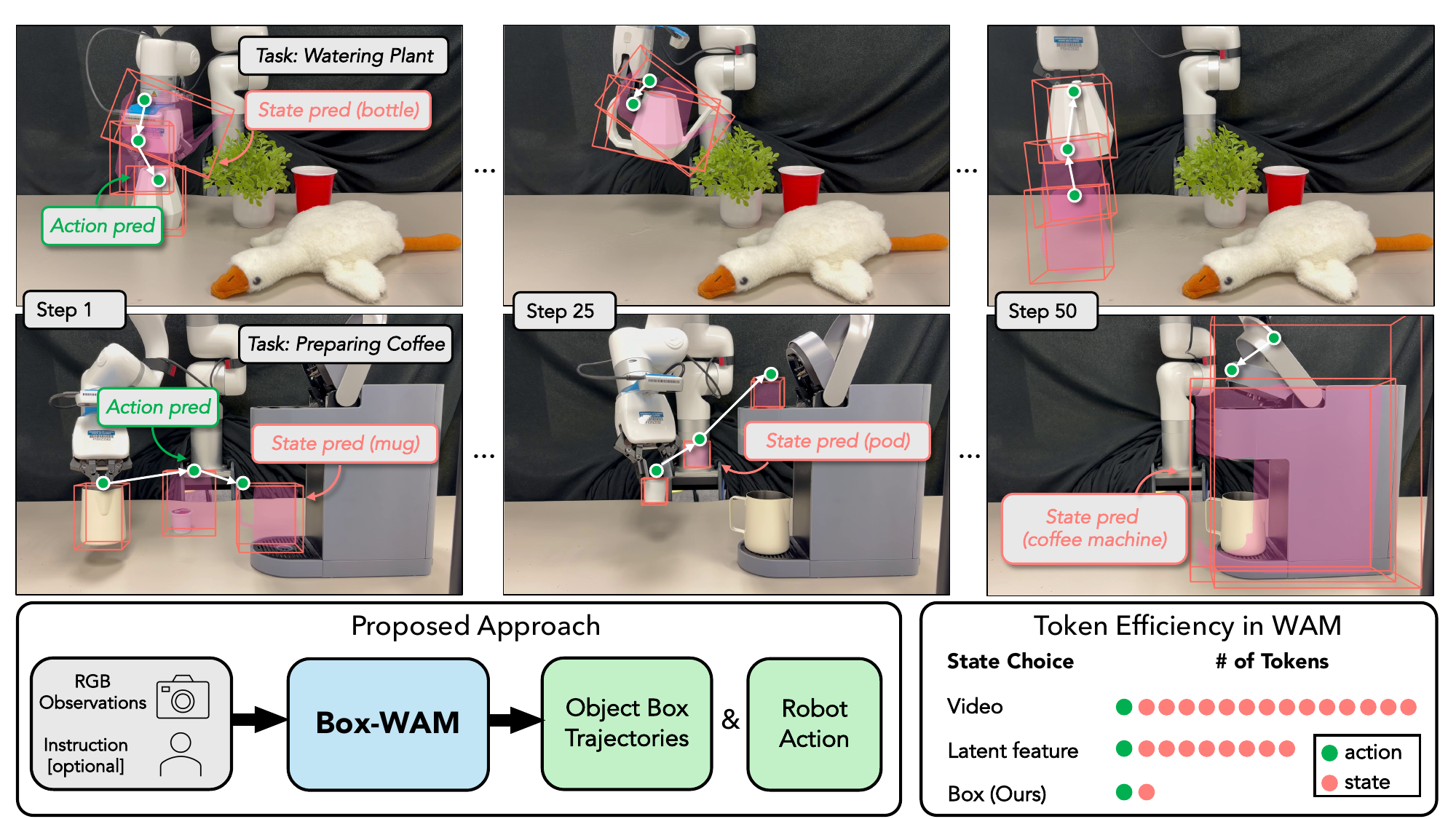}
\captionof{figure}{
\textbf{\methodname{} predicts compact object-centric futures for action generation.}
Instead of predicting dense future video, \methodname{} predicts future box trajectories of task-relevant objects and uses them to guide action generation. The predicted box automatically shifts to the relevant object at different stages of manipulation as shown in the coffee-making task.
}
\label{fig:teaser}
\end{center}

\begin{abstract}
World action models (WAMs) are a category of policy models that explicitly predict both actions and future world states. They have been shown to outperform action-only models across various robotic manipulation tasks. However, representing the future with full-frame video prediction can be computationally expensive during both training and inference. This raises a central question: what compact state representation preserves the planning and supervision benefits of future prediction? In this paper, we propose BoxWAM, which replaces dense future frames with oriented bounding box trajectories of task-relevant objects. By explicitly imagining object-centric future motion, our method provides a lightweight prediction target that captures the key physical changes induced by robot actions. We show that box trajectory prediction substantially reduces computational cost while improving action prediction performance, suggesting that object-centric compact states can serve as an efficient alternative to video-based world action modeling for robotic manipulation. Result videos can be found on https://box-wam.github.io
\end{abstract}

\section{Introduction}
\label{sec:introduction}
Behavior cloning (BC) is a simple and effective paradigm for robot manipulation, typically learning action-only policies that map observations directly to robot actions~\cite{black2024pi_0, black2025pi_, chi2025diffusion, kim2025openvla, ze20243d, zhao2023learning, rt12022arxiv, rt22023arxiv, octo_2023, liu2024rdt, kim24openvla, yang2025egovlalearningvisionlanguageactionmodels}.
Recently, however, world action models (WAMs) have emerged as a potentially superior paradigm for policy learning. WAMs take in observations and predict both actions and future environment states. Since the future state serves as an intermediate representation that guides action generation, WAMs have the potential to outperform action-only models in many manipulation tasks~\cite{bi2025motus, kim2026cosmos, ye2026world, yuan2026fastwam, lingbot-va2026}.

Although the approach is promising overall, the choice of future state representation for WAMs remains a key open question. Prior works have explored several options, including future video frames~\cite{bharadhwaj2024gen2act, liang2025video, li2025unified, liao2025genie, chen2025largevideoplanner, Tamar-RSS-19}, semantic masks~\cite{lou2026mask}, patch tokens, and latent features~\cite{rt12022arxiv, assran2025vjepa2, terver2025drivessuccessphysicalplanning, leworldmodel2026}. Dense representations are expressive but expensive: they require the model to predict background appearance, lighting, texture, and other visual details that are often irrelevant to the manipulation policy. Latent representations reduce this burden, but they can be difficult to learn reliably and are not interpretable. This creates a fundamental design question for WAMs: what kind of future state should a robot imagine in order to improve action prediction?

We study compact state representations that benefit WAMs. In this paper, we introduce BoxWAM, a world action model that predicts compact object-centric box trajectories and uses them to guide robot action prediction. The task-relevant object state is represented by the image-plane coordinates of the eight corners of an oriented 3D bounding box for each manipulated object in the scene. BoxWAM jointly denoises future box and action tokens, using a structured attention mask so actions can condition on imagined object motion without letting box predictions depend on future actions. The resulting box trajectory serves as a lightweight and interpretable object-motion plan guiding action generation.

We evaluate BoxWAM in simulation and the real world. In simulation, we use RoboTwin2.0~\cite{chen2025robotwin} to test task-specific imitation and multi-task generalist policy learning across diverse bimanual manipulation tasks. We also evaluate on three challenging real-world single-arm tasks. We compare against strong action-only policies and existing WAMs, and ablate the state representation and state-action coupling design. Our results support the hypothesis that box trajectories provide a compact yet effective future-state representation for world action models.

Our contributions are:
\begin{itemize}
    \item We propose box trajectories as an interpretable, object-centric future state representation for WAMs.
    \item We introduce BoxWAM, a rectified-flow transformer that jointly generates future box trajectories and actions with state-guided attention.
    \item We evaluate BoxWAM in simulation and on real robots, showing improved action prediction with lower future-modeling cost.
\end{itemize}

Code and videos are available at \url{https://box-wam.github.io/}.

\section{Related Work}
\label{sec:related}
\paragraph{World Models and World Action Models}
{\it World models} predict how an environment evolves over time and have been widely used for planning, policy learning, and representation learning in embodied agents~\citep{ha2018worldmodels, hafner2019dreamer, hafner2020dreamerv2, hafner2025dreamerv3, hafner2025training, alonso2024diffusionworldmodelingvisual, lee2026tracegen}.
In robotics, action-conditioned world models
~\citep{finn2017deep, visualforesight, yang2023learning, zhou2024dinowmworldmodelspretrained, lou2026mask, assran2025vjepa2, leworldmodel2026, nvidia2025cosmosworldfoundationmodel, nvidia2025cosmospredict2}
typically predict future states conditioned on actions, after which a planner selects an action sequence according to task objectives. This formulation is powerful but can require expensive action sampling or optimization at test time.
{\it World action models}~\citep{bi2025motus, kim2026cosmos, ye2026world, yuan2026fastwam, lingbot-va2026} instead predict actions and future states jointly. This shifts the role of future prediction from evaluating candidate actions to providing an intermediate structure for action generation. A key design question is therefore the choice of future representation. Prior work has considered video prediction~\citep{bharadhwaj2024gen2act, liang2025video, li2025unified, liao2025genie, chen2025largevideoplanner, Tamar-RSS-19}, semantic masks~\citep{lou2026mask}, learned latent states~\citep{rt12022arxiv, assran2025vjepa2, terver2025drivessuccessphysicalplanning, leworldmodel2026}, and tokenized visual futures. Another design choice is how state and action interact: state-then-action prediction ~\citep{kim2026cosmos}, joint prediction ~\citep{bi2025motus, ye2026world}, or auxiliary/surrogate state prediction ~\citep{yuan2026fastwam}.
{\it Our work} proposes a lightweight WAM design based on explicit object-centric box trajectories, and shows that imagining desired future box trajectories can improve action prediction.

\paragraph{Structured Representations for Robot Learning}

We are not the first to study compact {\it structured representations} for robot learning. In autonomous driving, motion forecasting commonly represents future scene evolution through object trajectories, including position, heading, and bounding-box-like agent states~\citep{Zeng_2019_CVPR, Zeng2020, Bansal_2019_RSS, hu2023_uniad}. In robotic manipulation, prior work~\citep{finn2016deep, manuelli2022kpam, qin2019keto, pmlr-v87-florence18a, simeonovdu2021ndf, zeng2020transporter, Zhang_2025_BMVC, ferraro2025focus, zhu2023learning, shridhar2022peract, shridhar2021cliport, huang2024rekep, Mo_2021_ICCV} has explored structured representations such as keypoints, including functional correspondences~\citep{tang2025mimicfunc}, object boxes, sparse motion fields, object-centric features, and dynamic scene representations. However, these representations are typically used in modular pipelines, such as model predictive control (MPC), inverse kinematics (IK), or as task-conditioning inputs, rather than as learned future states inside world action models. Our work revisits structured representations in the WAM paradigm and studies whether box trajectories can serve as a compact, action-relevant future state representation for manipulation.

\section{Method}
\label{sec:method}
\begin{figure*}[t]
    \centering
    \includegraphics[width=\textwidth,trim=0 7.5cm 0 0,]{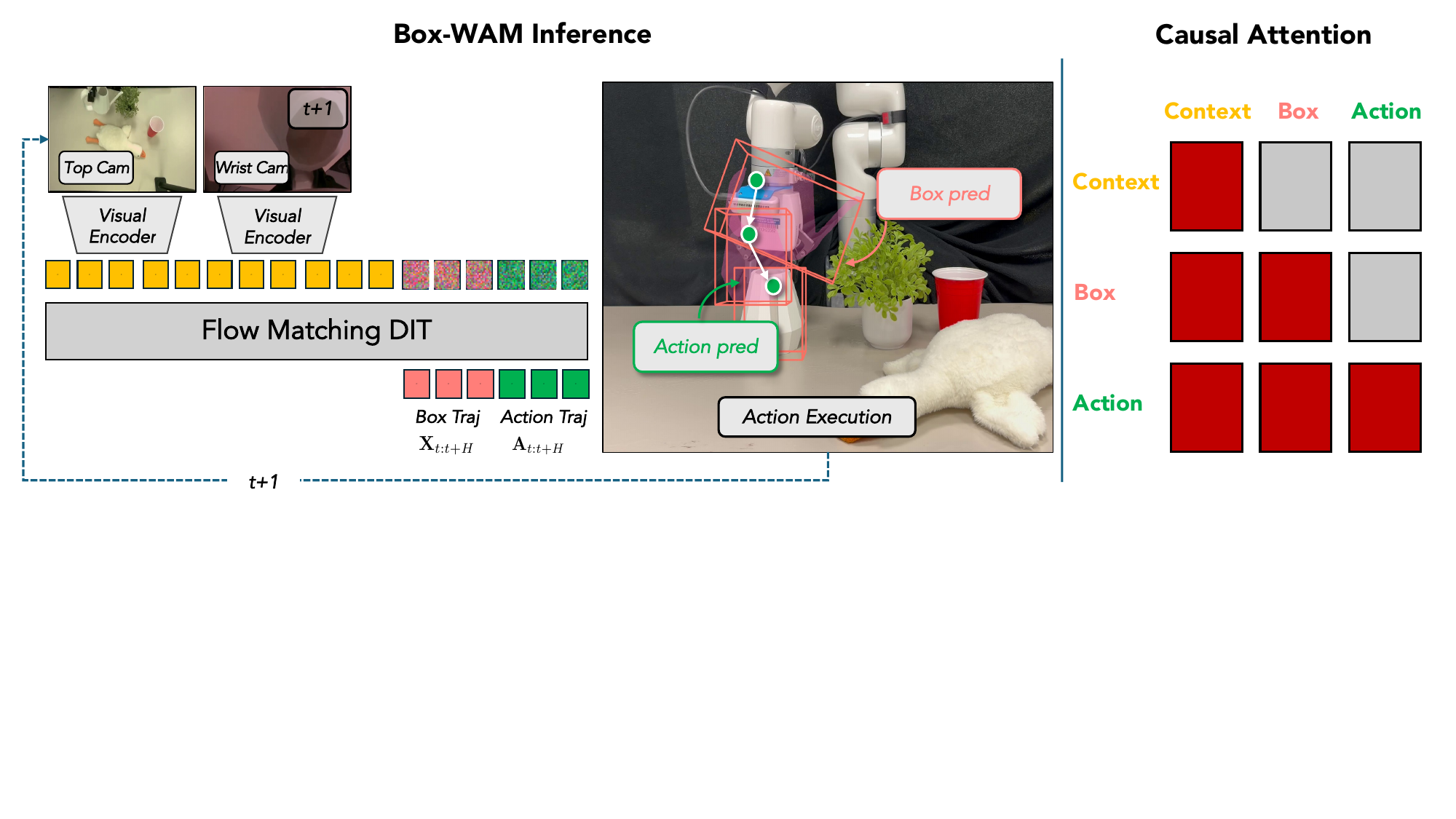}
    \vspace{-1em}
    \caption{
    \textbf{Overview of \methodname{}.}
    Observation images are encoded into visual context tokens, which condition a rectified-flow transformer that jointly denoises future state tokens and action tokens.
    An enforced causal attention allows action tokens to condition on imagined object motion while preventing future state tokens from depending on future action tokens.
    The decoded future state forms a compact box trajectory, and the decoded action tokens form the robot action chunk.
    }
    \label{fig:architecture}
    \vspace{-0.8em}
\end{figure*}

We present \methodname, a world action model that jointly predicts future task-centric box trajectories and robot actions. The key idea is to replace dense future observation prediction with a compact representation that captures task-relevant object motion. Fig.~\ref{fig:architecture} provides an overview of the architecture and the state-action interaction design.

\subsection{Problem Formulation and Representation}
\paragraph{Problem Formulation}

Given image observation $\mathbf{I}_T$ at timestep $T$, together with the current proprioceptive state $\mathbf{q}_T$ and task condition $\ell$, the goal is to predict a future state-action trajectory over horizon $H$:
$
    \left\{ \mathbf{x}_{T+h}, \mathbf{a}_{T+h} \right\}_{h=1}^{H},
$
where $\mathbf{x}_{T+h}$ denotes the future state representation and $\mathbf{a}_{T+h}$ denotes the corresponding robot action.
We model the joint conditional distribution
\begin{equation}
    p_\theta\!\left(
    \mathbf{x}_{T+1:T+H}, \mathbf{a}_{T+1:T+H}
    \mid \mathbf{I}_T, \mathbf{q}_T, \ell
    \right).
\end{equation}
The task condition $\ell$ is empty for single-task policies and is a language instruction for multi-task policies, at both training and inference.

\paragraph{Structured state representation}

We represent the future state $\mathbf{x}_{T+1:T+H}$ as an object-centric box trajectory for the \textit{actively manipulated} objects (explained later).
For each object, we use its minimum-volume oriented 3D bounding box, whose axes are chosen to tightly enclose the object geometry. At each time $t$, the state is defined as the image-plane coordinates of the eight corners of the oriented 
3D bounding box  projected into a third-person camera view:
$
    \mathbf{x}_t =
    \left[
    u_{1,t}, v_{1,t},
    \ldots,
    u_{8,t}, v_{8,t}
    \right] \in \mathbb{R}^{16}.
$
This representation preserves the coarse shape and motion of the manipulated object while discarding pixel-level details such as background texture, lighting variation, and shadows. 

For training, the active object is the currently manipulated object or, if none is being manipulated, the next object contacted in the demonstration. This uses future contact information to construct supervision targets only. At inference, the policy receives RGB observations, proprioception, and the multi-task language instruction; it receives neither ground-truth boxes nor future contact order, and predicts the active-object box trajectory itself. Figure~\ref{fig:active_object_random_order} shows different orders learned from randomized-order demonstrations.
In bimanual settings, we maintain separate box slots for the left and right arms:
$
    \mathbf{x}_t =
    \left[
    \mathbf{x}_t^{\mathrm{left}},
    \mathbf{x}_t^{\mathrm{right}}
    \right].
$
If an arm is not associated with an active object at timestep $t$, the corresponding slot is set to all zeros and masked out in the state-prediction loss.
In simulation, ground-truth boxes can be extracted directly from the simulator state. For real-world data, box supervision is reconstructed from RGB-D recordings, known meshes, and camera calibration, with rigid pose propagation during grasping (Appendix~\ref{sec:supp_real_obb_gt}). These additional inputs are used for annotation, not policy inference.

\paragraph{Action Representation}

The action sequence $\mathbf{a}_{T+1:T+H}$ is predicted over the same horizon as the box trajectory.
In the task-specific and 2,500-demonstration multi-task experiments, we use joint-position control. The task-specific policy represents each action as a relative joint-position command:
$
    \mathbf{a}_t = \mathbf{q}_t^{\mathrm{target}} - \mathbf{q}_t^{\mathrm{curr}}
$,
where $\mathbf{q}_t^{\mathrm{curr}}$ is the current absolute joint positions and $\mathbf{q}_t^{\mathrm{target}}$ is the desired target absolute joint positions.
During training, actions are normalized to $[-1, 1]$ and the model predicts normalized relative actions.
At inference time, the predicted action is denormalized and added back to the current joint configuration to recover an absolute target joint position:
$
    \mathbf{q}_t^{\mathrm{target}} = \mathbf{q}_t^{\mathrm{curr}}
     + \mathrm{denorm}(\mathbf{a}_t).
$
The resulting target joint positions are sent to the controller. 

\subsection{BoxWAM}
\paragraph{Task-Specific Model}

\methodname{} uses a transformer-based rectified-flow policy to jointly generate future box trajectories and actions.
First, observation images are encoded into visual context tokens using a frozen DINOv3~\citep{simeoni2025dinov3} vision encoder.
The model then concatenates visual context tokens, noisy future box tokens, and noisy action tokens into a single transformer sequence.
For each future timestep, we allocate one box token and one action token.
Thus, for horizon $H$, the model denoises $H$ box tokens and $H$ action tokens.
The transformer backbone processes the full token sequence, while the current proprioceptive state and flow timestep are injected into each block through adaptive layer normalization.
After denoising, a linear future-state head maps box tokens to future box coordinates, and a linear action head maps action tokens to robot actions. 

\paragraph{State-Action Interaction}

A central design choice in \methodname{} is how future state tokens and action tokens interact during generation. Although future states and actions are predicted within a unified flow-matching model, we enforce a causal state-action attention structure: action tokens can attend to future state tokens, while future state tokens cannot attend to future action tokens. Box tokens attend to visual context and other box tokens; action tokens attend to visual context, box tokens, and other action tokens. This design allows the predicted future box trajectory to serve as a task-centric object-motion plan that guides action generation.

\paragraph{Training Objective}

We train \methodname{} with a rectified-flow objective over the joint future variable
$
    \mathbf{z}^1 = [\mathbf{x}^1_{T+1:T+H}, \mathbf{a}^1_{T+1:T+H}],
$
where $\mathbf{x}^1$ is the ground-truth future box trajectory and $\mathbf{a}^1$ is the ground-truth action chunk.
Given Gaussian noise $\boldsymbol{\epsilon} \sim \mathcal{N}(0, I)$ and a flow timestep $\tau \in [0,1]$, we construct an interpolated sample
$
    \mathbf{z}^\tau = (1-\tau)\boldsymbol{\epsilon} + \tau \mathbf{z}^1.
$
The model predicts the velocity field from noise to data.
Since $\mathbf{z}$ contains both box and action variables, we decompose the predicted velocity into state and action components:
$
    \hat{v}_\theta(\mathbf{z}^\tau, \tau)
    =
    \left[
    \hat{v}^X_\theta(\mathbf{z}^\tau, \tau),
    \hat{v}^A_\theta(\mathbf{z}^\tau, \tau)
    \right].
$
The training loss is
\begin{equation}
\mathcal{L}
=
\mathbb{E}_{\tau, \boldsymbol{\epsilon}_A, \boldsymbol{\epsilon}_X, \mathbf{z}^1} \left[
\lambda_X
\left\|
\hat{v}^{X}_\theta(\mathbf{z}^\tau, \tau)
-
(\mathbf{x}^1 - \boldsymbol{\epsilon}_X)
\right\|_2^2
+
\lambda_A
\left\|
\hat{v}^{A}_\theta(\mathbf{z}^\tau, \tau)
-
(\mathbf{a}^1 - \boldsymbol{\epsilon}_A)
\right\|_2^2 \right] ,
\end{equation}
where $\lambda_X$ and $\lambda_A$ balance future-state and action prediction. We use equal weights for the two losses ($\lambda_X=\lambda_A$).

Following the work of $\pi_{0.5}$~\citep{black2025pi_}, we sample flow timesteps based on the Beta distribution.
Specifically, we draw
$
    u \sim \mathrm{Beta}(1.5, 1.0),
$
and set
$
    \tau = 0.999(1-u).
$
This biases training toward smaller $\tau$, exposing the model more frequently to noisier points along the rectified-flow trajectory.

\paragraph{Inference}
At test time, \methodname{} samples initial Gaussian noise for both future box tokens and action tokens.
The model then integrates the learned rectified-flow ordinary differential equation from $\tau=0$ to $\tau=1$ using Euler integration:
$
    \mathbf{z}^{\tau_{k+1}}
    =
    \mathbf{z}^{\tau_k}
    +
    (\tau_{k+1} - \tau_k)
    \hat{v}_\theta(\mathbf{z}^{\tau_k}, \tau_k).
$
For $N$ denoising steps, we use a quadratic integration schedule
$
    \tau_k =
    1 -
    \left(
    1 - \frac{k}{N}
    \right)^2,
    \quad k=0,\ldots,N
$
which allocates more integration steps near $\tau=1$.
Box tokens and action tokens are updated together at each denoising step, and no additional noise is injected after initialization.

After integration, the denoised action tokens are decoded into an action chunk.
The policy executes the full action chunk before replanning, except in the 27,500-demonstration comparison, which executes 36 actions. 
We use a prediction horizon of $H=15$ for task-specific simulation experiments and $H=50$ for multi-task and real-world settings.

\paragraph{Integration with $\pi_{0.5}$}
We add 50 future-box tokens and two linear projection layers to the $\pi_{0.5}$ action expert, using the same state-action attention mask. Both policies in Table~\ref{tab:generalization} are fully fine-tuned from the same checkpoint on 2,500 demonstrations for 30,000 steps; BoxWAM additionally receives box-trajectory supervision.

\section{Experiments}
\label{sec:result}
\begin{figure*}[htbp]
    \centering
    \setlength{\fboxsep}{0pt}
    \setlength{\fboxrule}{0.3pt}

    \newcommand{\rolloutfig}[2]{%
        \begin{minipage}[t]{0.25\textwidth}
            \centering
            \fbox{\includegraphics[width=0.96\linewidth,height=0.13\textheight,keepaspectratio]{#1}}
            \\[0.25em]
            \textbf{#2}
        \end{minipage}%
    }

    \rolloutfig{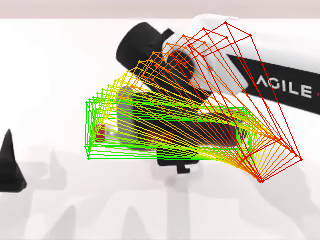}{\small{adjust bottle}}%
    \rolloutfig{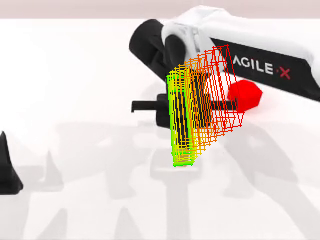}{\small{beat block hammer}}%
    \rolloutfig{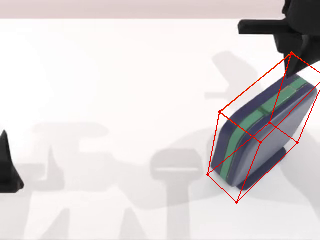}{\small{click alarmclock}}%
    \rolloutfig{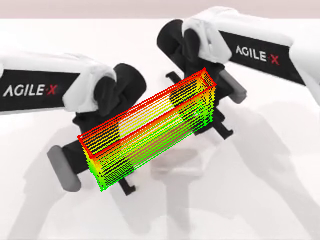}{\small{grab roller}}

    \vspace{0.6em}

    \rolloutfig{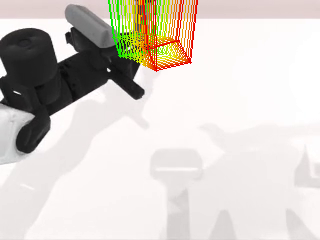}{\small{handover mic}}%
    \rolloutfig{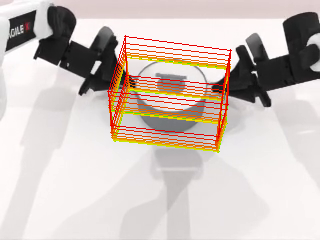}{\small{lift pot}}%
    \rolloutfig{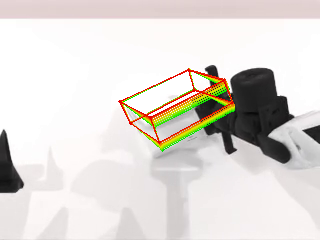}{\small{open laptop}}%
    \rolloutfig{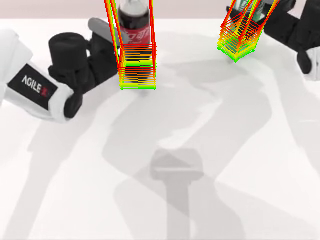}{\small{pick dual bottles}}

    \caption{
    \textbf{Simulation rollout visualizations on RoboTwin2.0.}
    We visualize representative rollouts from eight simulation tasks with predicted future box trajectories overlaid on the scene.
    \methodname{} predicts compact object-centric future motion across diverse manipulation behaviors.
    }
    \label{fig:sim_rollouts}
    \vspace{-0.8em}
\end{figure*}

Our experiments answer three questions about \methodname{}: 
{\bf 1)} {\it Does \methodname{} outperform strong action-only policies? }
{\bf 2)} {\it Are box trajectories more effective and efficient than alternative future-state representations in WAMs? }
{\bf 3)} {\it Can box-trajectory prediction serve as a WAM plugin to improve pretrained generalizable VLA models?}

\subsection{Experimental Setup}
\paragraph{Task-Specific Simulation Benchmark}
We evaluate \methodname{} on RoboTwin2.0~\citep{chen2025robotwin}, a simulation benchmark covering diverse robotic manipulation tasks that require object interaction, spatial reasoning, long-horizon control, and bimanual coordination. 
The benchmark provides paired observations, future states, and action trajectories, allowing us to supervise both future-state prediction and action prediction. 
RoboTwin2.0 includes a clean setting with object-pose randomization and a randomized setting with more challenging scene-configuration variation. Models are trained on the clean setting data only but are evaluated on both settings to test robustness against scene changes. 
For task-specific policy comparisons, our models are trained with 50 demonstrations per task and evaluated over 3 random seeds, with 100 trials per seed. Other baseline entries in Table~\ref{tab:single_task_clean_randomized} are taken from the RoboTwin2.0 leaderboard. 
We compare \methodname{} against ACT~\citep{zhao2023learning}, Diffusion Policy~\citep{chi2025diffusion}, DP3~\citep{ze20243d}, $\pi_0$~\citep{black2024pi_0} with one checkpoint per task, and an action-only variant of our architecture, which effectively acts as flow-matching policy. 

\paragraph{Multi-Task Simulation Benchmark}

We further evaluate \methodname{} under a multi-task generalist setting. For Table~\ref{tab:generalization}, models are trained on a multi-task dataset consisting of 50 tasks and 50 demonstrations per task, resulting in a total of 2500 demonstrations.  This setting measures whether \methodname{} scales beyond task-specific training. For each task, we evaluate over 100 trials on one seed.
We compare against pretrained VLA-style baselines: $\pi_{0.5}$ and a BoxWAM-augmented $\pi_{0.5}$ that predicts future box trajectories with the same state-action attention design.
 Evaluations are conducted on both a clean setting that follows the same setup as data collection, and a randomized setting with randomized scene configurations (RoboTwin2.0 Hard~\citep{chen2025robotwin}) to test whether compact box trajectories provide a useful future representation under distribution shift.

\paragraph{Real-World Benchmark}

We deploy \methodname{} on a real robot with a single-arm UFACTORY xArm7 setup, a third-person scene camera that observes the full workspace, and a wrist-mounted camera. Training data is collected from human demonstrations on a GELLO~\citep{wu2024gello} teleoperation system. We evaluate three real-world manipulation tasks: making coffee, watering a plant, and placing a Pokemon toy onto a plate. We compare \methodname{} with Diffusion Policy and an action-only variant of our model under the same execution protocol. For each task, we collect 50 demonstrations and evaluate on 50 randomized test scenarios. Because full task success can be sparse, we report both final success rate and progress rate over task-specific subgoals (Appendix~\ref{sec:supp_real_progress}). 

\subsection{Results}

\paragraph{Task-Specific Simulation Results}

\begin{table*}[t]
\centering
\footnotesize
\caption{\textbf{Single-task RoboTwin2.0 success rates (\%).} BoxWAM and Flow (action-only) are evaluated over three seeds with 100 trials per seed for each task and setting. Other baselines are from the RoboTwin2.0 leaderboard. Gray rows indicate pretrained models.}
\label{tab:single_task_clean_randomized}

\setlength{\tabcolsep}{1.9pt}
\renewcommand{\arraystretch}{1.12}

\begin{tabular*}{\textwidth}{@{\extracolsep{\fill}} l c c c c c c c c c c c}
\toprule
\textbf{Method} &
\textbf{Obs.} &
\textbf{Avg.} &
\makecell{\textbf{Adjust}\\ \textbf{Bottle}} &
\makecell{\textbf{Beat}\\ \textbf{Hammer}} &
\makecell{\textbf{Click}\\ \textbf{Alarm}} &
\makecell{\textbf{Grab}\\ \textbf{Roller}} &
\makecell{\textbf{Handover}\\ \textbf{Mic}} &
\makecell{\textbf{Lift}\\ \textbf{Pot}} &
\makecell{\textbf{Open}\\ \textbf{Laptop}} &
\makecell{\textbf{Pick Dual}\\ \textbf{Bottles}} &
\makecell{\textbf{Turn}\\ \textbf{Switch}} \\
\midrule
\multicolumn{12}{c}{\textbf{Clean setting}} \\
\midrule
DP & 1 RGB & 55.4 & 97.0 & 42.0 & 61.0 & 98.0 & 53.0 & 39.0 & 49.0 & 24.0 & 36.0 \\
ACT & 3 RGB & 60.4 & 97.0 & 56.0 & 32.0 & 94.0 & 85.0 & \textcolor{blue}{\textbf{88.0}} & 56.0 & 31.0 & 5.0 \\
DP3 & 1 RGBD & \textbf{81.2} & \textcolor{blue}{\textbf{99.0}} & \textbf{72.0} & 77.0 & 98.0 & \textbf{100.0} & \textbf{97.0} & \textcolor{blue}{\textbf{82.0}} & \textcolor{blue}{\textbf{60.0}} & 46.0 \\
\textcolor{gray}{$\pi_0$} &
\textcolor{gray}{3 RGB} &
\textcolor{gray}{71.4} &
\textcolor{gray}{90.0} &
\textcolor{gray}{43.0} &
\textcolor{gray}{63.0} &
\textcolor{gray}{96.0} &
\textcolor{gray}{\textbf{98.0}} &
\textcolor{gray}{84.0} &
\textcolor{gray}{\textbf{85.0}} &
\textcolor{gray}{57.0} &
\textcolor{gray}{27.0} \\
\makecell[l]{Flow\\(action-only)} & 1 RGB & 73.2 & \textbf{99.7} & 63.0 & \textcolor{blue}{\textbf{90.3}} & \textcolor{blue}{\textbf{99.7}} & 58.3 & 69.3 & 81.0 & 47.7 & \textcolor{blue}{\textbf{50.0}} \\
\methodname{} & 1 RGB & \textcolor{blue}{\textbf{78.3}} & 97.7 & \textcolor{blue}{\textbf{67.7}} & \textbf{95.0} & \textbf{100.0} & 65.3 & 84.3 & 80.0 & \textbf{62.0} & \textbf{52.3} \\
\midrule
\multicolumn{12}{c}{\textbf{Randomized setting}} \\
\midrule
DP & 1 RGB & 0.7 & 0.0 & 0.0 & 5.0 & 0.0 & 0.0 & 0.0 & 0.0 & 0.0 & 1.0 \\
ACT & 3 RGB & 6.3 & 23.0 & 3.0 & 4.0 & 25.0 & 0.0 & 0.0 & 0.0 & 0.0 & 2.0 \\
DP3 & 1 RGBD & 5.1 & 3.0 & 8.0 & 14.0 & 2.0 & \textcolor{blue}{\textbf{3.0}} & 0.0 & 7.0 & 1.0 & 8.0 \\
\textcolor{gray}{$\pi_0$} &
\textcolor{gray}{3 RGB} &
\textcolor{gray}{\textbf{33.1}} &
\textcolor{gray}{56.0} &
\textcolor{gray}{\textbf{21.0}} &
\textcolor{gray}{11.0} &
\textcolor{gray}{\textbf{80.0}} &
\textcolor{gray}{\textbf{13.0}} &
\textcolor{gray}{\textbf{36.0}} &
\textcolor{gray}{\textbf{46.0}} &
\textcolor{gray}{12.0} &
\textcolor{gray}{23.0} \\
\makecell[l]{Flow\\(action-only)} & 1 RGB & 32.6 & \textcolor{blue}{\textbf{72.3}} & 18.7 & \textcolor{blue}{\textbf{17.7}} & 77.7 & 1.0 & 6.3 & \textcolor{blue}{\textbf{43.3}} & \textbf{18.3} & \textbf{38.0} \\
\methodname{} & 1 RGB & \textbf{37.8} & \textbf{88.7} & \textbf{29.3} & \textbf{29.3} & \textcolor{blue}{\textbf{78.3}} & 0.0 & \textcolor{blue}{\textbf{22.7}} & 40.3 & \textcolor{blue}{\textbf{16.7}} & \textcolor{blue}{\textbf{34.7}} \\
\bottomrule
\end{tabular*}
\end{table*}

\begin{figure*}[t]
    \centering
    \small
    \setlength{\tabcolsep}{0pt} %

    \begin{tabular}{@{}c@{}c@{}c@{}c@{}}
        \includegraphics[width=0.249\textwidth]{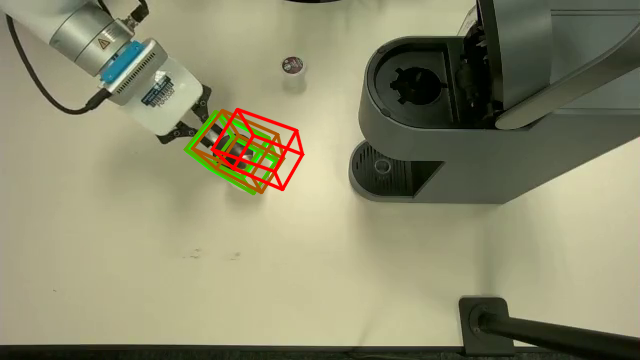} &
        \includegraphics[width=0.249\textwidth]{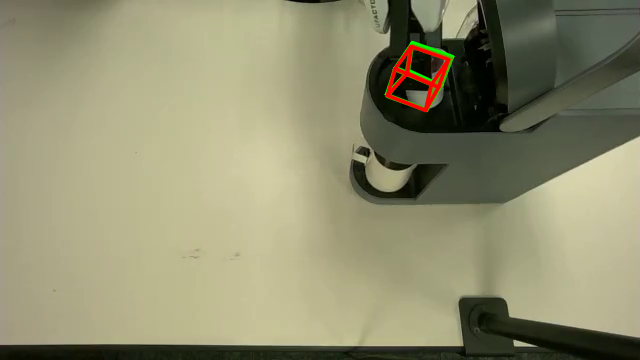} &
        \includegraphics[width=0.249\textwidth]{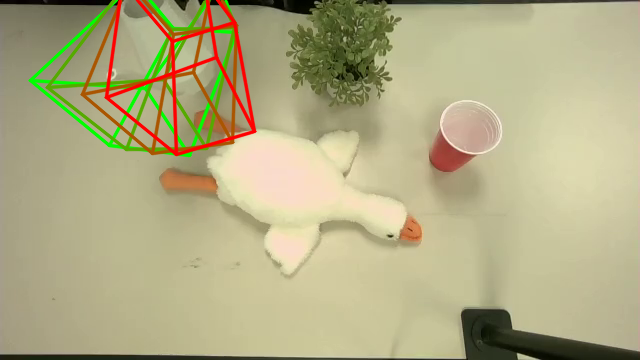} &
        \includegraphics[width=0.249\textwidth]{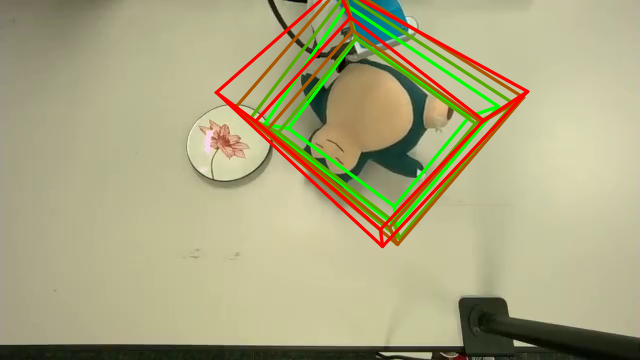} \\[-0.2em]

        \includegraphics[width=0.249\textwidth]{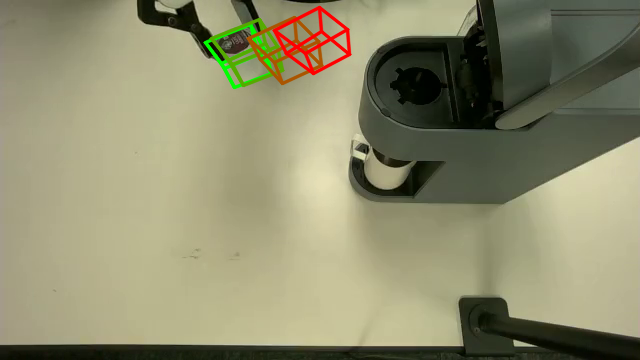} &
        \includegraphics[width=0.249\textwidth]{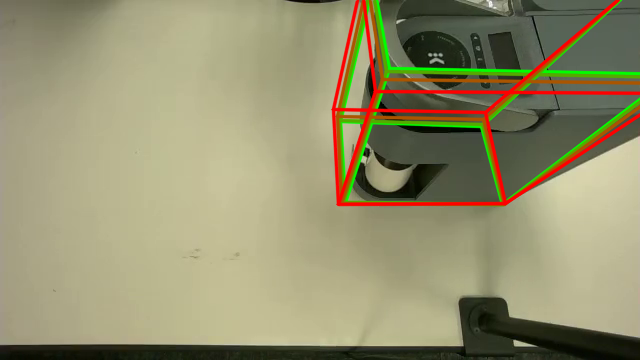} &
        \includegraphics[width=0.249\textwidth]{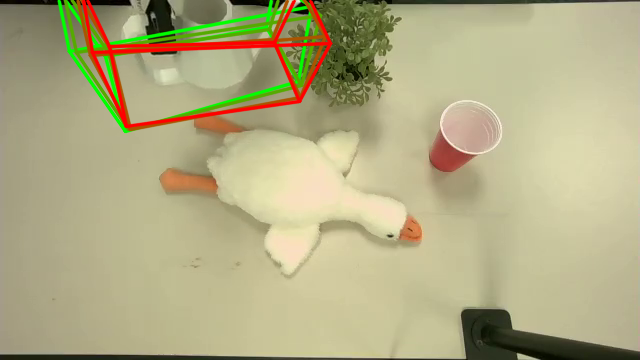} &
        \includegraphics[width=0.249\textwidth]{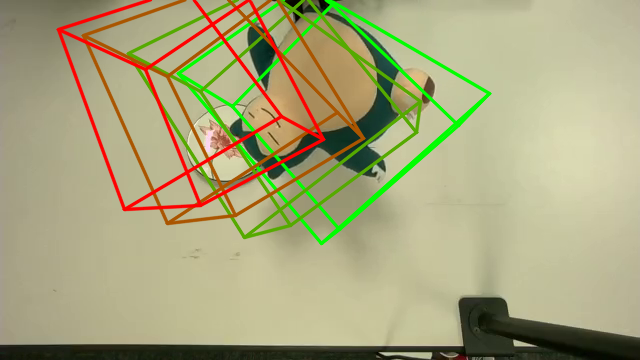} \\[0.25em]

        \textbf{Make Coffee (Early)} &
        \textbf{Make Coffee (Late)} &
        \textbf{Water Plant} &
        \textbf{Place Pokemon}
    \end{tabular}

    \caption{
    \textbf{Real-world rollout visualizations.}
    We visualize representative rollouts from three real-world manipulation tasks with predicted future box trajectories overlaid on the scene-camera view.
    The first two columns show early and late coffee-making stages; the remaining columns show watering and toy placement. Each row shows a different frame, illustrating how \methodname{} predicts compact object-motion plans for action generation under real-world perception noise, actuation error, and object pose variation.}
    \label{fig:real_world_rollouts}
    \vspace{-0.2em}
\end{figure*}

Table~\ref{tab:single_task_clean_randomized} shows that BoxWAM exhibits performance gains on tasks that require object motion reasoning, spatial alignment and long-horizon manipulation. In the clean setting, it on average outperforms all baselines except DP3 (which relies on depth input). In the randomized setting, it achieves the highest average success rate.
Figure~\ref{fig:sim_rollouts} visualizes selected simulation rollouts together with predicted box trajectories.

\begin{table}[t]
\centering
\small
\caption{
\textbf{Multi-task fine-tuning} with 2,500 demonstrations and joint-position control. Both models are trained for 30,000 steps for a fair comparison.
Models are evaluated under clean and randomized settings.
We report success rate (\%). The inference latency (ms) is reported based on default 10 denoising steps from $\pi_{0.5}$.
}
\label{tab:generalization}
\begin{tabular}{l c c c}
\toprule
\textbf{Method} &
\textbf{Clean} &
\textbf{Randomized} &
\textbf{Latency} \\
\midrule
$\pi_{0.5}$ & 52.6 & 26.9 & \textbf{55.3} \\
\textbf{$\pi_{0.5}$ + \methodname{}} & \textbf{61.5} & \textbf{34.0} & 57.7 \\
\bottomrule
\end{tabular}
\vspace{-.5em}
\end{table}

\paragraph{Multi-Task Simulation Results}
We next study whether box trajectories remain useful in a multi-task  setting. Table~\ref{tab:generalization} reports results for $\pi_{0.5}$ and a BoxWAM-augmented variant trained across multiple RoboTwin2.0 tasks. Adding box-trajectory prediction improves clean performance from $52.6\%$ to $61.5\%$, and randomized performance from $26.9\%$ to $34.0\%$. These gains support the utility of additional box supervision and prediction for pretrained VLAs under matched data and training steps. Improvements are not uniform across tasks; Appendix~\ref{sec:supp_multitask} reports the full breakdown.

\begin{table*}[t]
\centering
\small
\caption{\textbf{Real-world manipulation results.} We report both final success rate and progress rate because full task success can be sparse. Each method is evaluated under the same execution protocol and randomized object pose.}
\label{tab:real_world}
{
\begin{tabular}{l c c c c c c c}
\toprule
\textbf{Method} &
\multicolumn{2}{c}{\textbf{Make Coffee}} &
\multicolumn{2}{c}{\textbf{Water Plant}} &
\multicolumn{2}{c}{\textbf{Place Pokemon}} &
\textbf{Average} \\
\cmidrule(lr){2-3}
\cmidrule(lr){4-5}
\cmidrule(lr){6-7}
&
\textbf{SR} &
\textbf{Progress} &
\textbf{SR} &
\textbf{Progress} &
\textbf{SR} &
\textbf{Progress} &
\\
\midrule
Diffusion Policy & 10.0 & 34.0 & 8.0 & 9.4 & 34.0 & 47.0 & 23.7 \\
Flow (action-only) & \textcolor{blue}{\textbf{24.0}} & \textcolor{blue}{\textbf{48.0}} & \textcolor{blue}{\textbf{48.0}} & \textcolor{blue}{\textbf{54.7}} & \textcolor{blue}{\textbf{60.0}} & \textcolor{blue}{\textbf{72.0}} & \textcolor{blue}{\textbf{51.1}} \\
\textbf{\methodname{}} &
\textbf{26.0} & \textbf{54.4} &
\textbf{66.0} & \textbf{70.7} &
\textbf{70.0} & \textbf{78.0} &
\textbf{60.9} \\
\bottomrule
\end{tabular}
}
\vspace{-.5em}
\end{table*}

\paragraph{Real-World Results} 
As shown in Table~\ref{tab:real_world}, BoxWAM outperforms both Diffusion Policy and the action-only variant under the same execution protocol. Averaged across success and progress metrics, BoxWAM achieves $60.85\%$, compared with $51.1\%$ for the action-only model and $23.7\%$ for Diffusion Policy. The strong performance of the action-only variant is likely due in part to the DINOv3 visual encoder, which provides stronger visual representations than the encoder used by the Diffusion Policy. However, BoxWAM further improves over this action-only model, showing that predicting future box trajectories provides additional object-centric structure that improves robustness during real-world execution.
Figure~\ref{fig:real_world_rollouts} shows that the predicted boxes form an interpretable object-motion plan that evolves with the task stage. These results suggest that compact future-state prediction transfers beyond simulation despite perception noise, actuation error, and randomized object configurations.

\subsection{Ablations and Runtime Analysis}
We evaluate ablations over 3 random seeds, with 50 trials per seed for each task and setting, and report the average success rate across nine tasks.
\begin{table}[t]
\centering
\begin{minipage}[t]{0.46\linewidth}
\vspace{0pt}
\caption{\textbf{Future-state ablations.} SR (\%); \\
"\textcolor{blue}{---}": same as box latency. }
\label{tab:state_prediction_ablation}
\label{tab:compact_states}
\end{minipage}\hfill
\begin{minipage}[t]{0.522\linewidth}
\vspace{0pt}
\caption{\textbf{Ablation on state-action interaction designs.} Arrows indicate information flow.}
\label{tab:state_action_ablation}
\end{minipage}
\par\noindent
\begin{minipage}[t]{0.46\linewidth}
\vspace{0pt}
\centering
\footnotesize
\setlength{\tabcolsep}{1.9pt}
\renewcommand{\arraystretch}{1}
\begin{tabular}{lrrr}
\toprule
\textbf{Representation} & \textbf{Latency} & \textbf{Clean} & \makecell{\textbf{Random-}\\\textbf{ized}} \\
\midrule
Flow (action-only) & \textbf{168.3 ms} & 73.2 & 32.6 \\
VAE & 250.5 ms & \textbf{80.1} & \textcolor{blue}{\textbf{34.5}} \\
DINO & 251.4 ms & 69.9 & 29.0 \\
6D object pose & \textcolor{blue}{---} & 73.2 & 32.1 \\
2D OBB & \textcolor{blue}{---} & 74.2 & 30.7 \\
Functional keypoints & \textcolor{blue}{---} & 75.3 & 27.8 \\
EE trajectory & \textcolor{blue}{---} & 74.0 & 32.6 \\
\textbf{Box (3D OBB)} & \textcolor{blue}{\textbf{172.4 ms}} & \textcolor{blue}{\textbf{78.3}} & \textbf{37.8} \\
\bottomrule
\end{tabular}
\end{minipage}\hfill
\begin{minipage}[t]{0.522\linewidth}
\vspace{0pt}
\centering
\footnotesize
\setlength{\tabcolsep}{2pt}
\begin{tabular}{lccc}
\toprule
\textbf{Design} & \makecell{\textbf{Attention}\\\textbf{Design}} & \textbf{Clean} & \makecell{\textbf{Random-}\\\textbf{ized}} \\
\midrule
Flow (action-only) & N/A & 73.2 & 32.6 \\
\makecell[l]{Fully separate\\prediction} & State $\perp$ Action & 76.6 & \textcolor{blue}{\textbf{37.4}} \\
Fully joint prediction & State $\Leftrightarrow$ Action & \textcolor{blue}{\textbf{76.7}} & 32.9 \\
Action then Future & State $\leftarrow$ Action & 76.1 & 35.5 \\
\makecell[l]{\textbf{Future then action}\\\textbf{(BoxWAM)}} & State $\rightarrow$ Action & \textbf{78.3} & \textbf{37.8} \\
\bottomrule
\end{tabular}
\end{minipage}
\end{table}

\paragraph{State Representation Ablation}
Table~\ref{tab:state_prediction_ablation} compares our box trajectories with an action-only baseline, Wan2.2 VAE~\citep{wan2025} tokens, DINOv3 ViT-L tokens, 6D object poses, 2D OBBs, functional keypoints, and end-effector trajectories over task-specific simulation benchmark. Functional keypoints~\citep{tang2025mimicfunc} use grasp, center, and function points, reusing grasp when function is absent.
All variants use the same action-prediction objective with different future-state targets and tokenization. We report task success for all variants and available latency measurements. The action-only and other variants are trained for 6{,}000 optimization steps.
Since the dense VAE and DINO future representations require substantially longer training to converge, we train them for 15{,}000 optimization steps.

Dense visual features do not consistently improve success despite introducing substantially more tokens. VAE achieves the highest clean success rate, while box trajectories achieve the highest randomized success rate and outperform the other compact states in both settings. Box trajectories also improve over the action-only baseline while adding only lightweight prediction overhead.

\begin{figure}[htbp]
\centering
\begin{minipage}[t]{0.48\linewidth}
\vspace{0pt}
\centering
\setlength{\tabcolsep}{1pt}
\begin{tabular}{ccc}
\small Rollout 1 & \small Rollout 2 & \small Rollout 3 \\
\includegraphics[width=.32\linewidth]{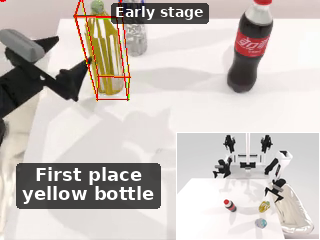} &
\includegraphics[width=.32\linewidth]{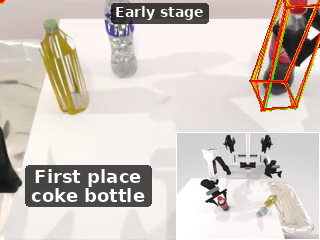} &
\includegraphics[width=.32\linewidth]{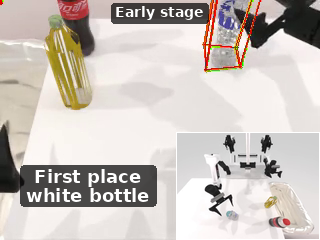} \\
\includegraphics[width=.32\linewidth]{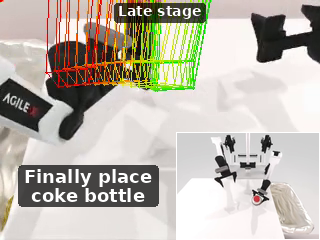} &
\includegraphics[width=.32\linewidth]{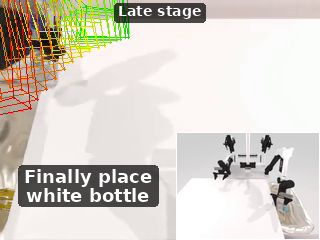} &
\includegraphics[width=.32\linewidth]{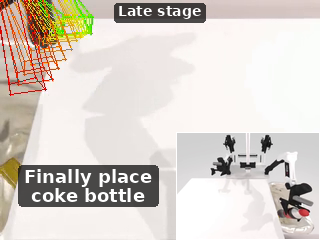}
\end{tabular}
\caption{\textbf{Multimodal ordering behavior} in Put Bottle Dustbin, learned from randomized-order demonstrations. Predicted boxes indicate the intended active object.}
\label{fig:active_object_random_order}
\end{minipage}\hfill
\begin{minipage}[t]{0.49\linewidth}
\vspace{0pt}
\centering
\includegraphics[width=\linewidth]{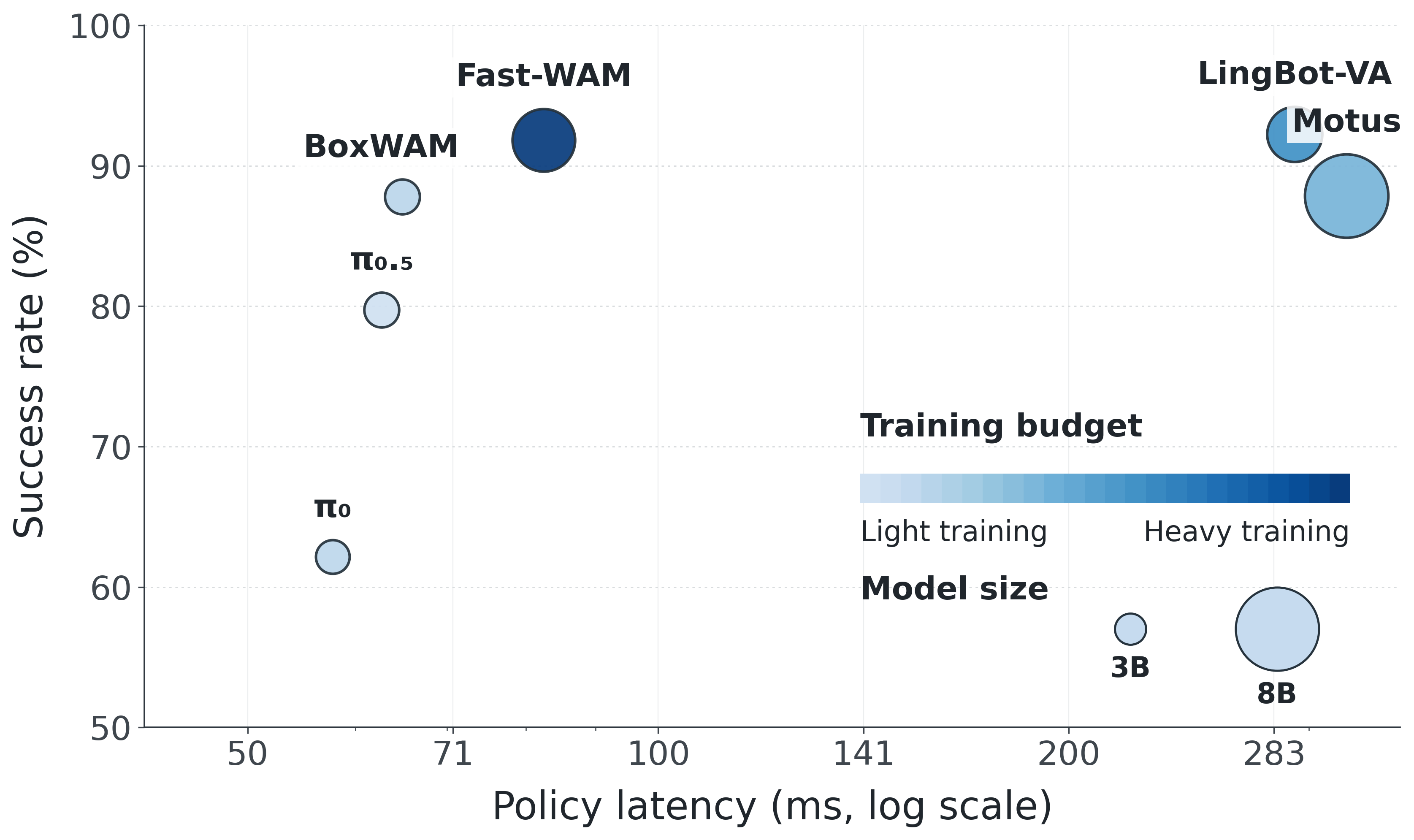}
\caption{\textbf{Comparison with existing models}. The vertical axis shows the mean success rate across the Easy and
Hard settings. Bubble size scales with model size, and darker blue indicates
a larger training budget in GPU-hours. Training-budget estimates are based on
reported configurations and runtimes.}
\label{fig:wam_efficiency}
\end{minipage}
\end{figure}

\paragraph{State-Action Interaction Ablation}
We ablate how future-state and action tokens interact inside the generative policy.
Table~\ref{tab:state_action_ablation} compares action-only prediction, fully separate prediction, fully joint prediction, action-then-future prediction, and our future-then-action design.
The results show that \methodname{} performs best when action tokens can attend to future state tokens, while future state tokens cannot attend to future action tokens, supporting our design choice. Competitive separate prediction also suggests a benefit from auxiliary supervision.

\paragraph{Runtime Analysis}
We report the runtime of \methodname{} and the baselines in Table~\ref{tab:generalization} and Table~\ref{tab:state_prediction_ablation}. All experiments are conducted on a RTX Pro 6000 Blackwell GPU. The runtime reflects how much computation is saved by replacing full video-style future prediction with compact box trajectories, and how much overhead \methodname{} adds compared with action-only inference. The results highlights the efficiency advantage of box trajectories as a lightweight future representation.

\paragraph{Comparison with Existing WAMs and an IK Baseline}
\label{sec:wam_comparison}

Using 27,500 demonstrations and end-effector control, we fine-tune $\pi_{0.5}$+BoxWAM for 35,000 steps, predicting 50 actions and executing 36. Figure~\ref{fig:wam_efficiency} compares success, latency (RTX Pro 6000), and model size, showing BoxWAM's performance--efficiency trade-off. While using a much smaller training budget, BoxWAM achieves 87.94\% SR in easy and 87.62\% SR in hard setting
at 64.9 ms with 3.35B parameters.

Differential IK using Mink~\citep{zakka2026mink} and M2T2~\citep{yuan2023m2t2}, with ground-truth task plans, poses, depth, and masks but no task-specific tuning, achieves 26.78\% Easy and 26.30\% Hard success.

\section{Conclusion}
\label{sec:conclusion}
We presented \methodname, a world action model that uses compact object-centric box trajectories as future states for action generation. By predicting the future motion of task-relevant objects instead of dense visual futures, \methodname{} provides an efficient and interpretable state representation for manipulation. Across simulation and real-world experiments, \methodname{} improves over action-only policies while adding minimal inference overhead. Our ablations show that box trajectories are most effective when used as intermediate object-motion plans that guide action prediction.

\section{Limitations}
\label{sec:limitations}
Our approach depends on reliable object tracking for ground-truth supervision and may degrade under severe occlusion, highly deformable objects, or tasks where boxes are insufficiently informative. However, our results overall suggest that compact object-centric futures are a promising interface for efficient world action modeling.

\clearpage
\bibliography{example}

\begin{thebibliography}{64}
\providecommand{\natexlab}[1]{#1}
\providecommand{\url}[1]{\texttt{#1}}
\expandafter\ifx\csname urlstyle\endcsname\relax
  \providecommand{\doi}[1]{doi: #1}\else
  \providecommand{\doi}{doi: \begingroup \urlstyle{rm}\Url}\fi

\bibitem[Black et~al.(2024)Black, Brown, Driess, Esmail, Equi, Finn, Fusai,
  Groom, Hausman, Ichter, et~al.]{black2024pi_0}
K.~Black, N.~Brown, D.~Driess, A.~Esmail, M.~Equi, C.~Finn, N.~Fusai, L.~Groom,
  K.~Hausman, B.~Ichter, et~al.
\newblock $\pi_0 $: A vision-language-action flow model for general robot
  control.
\newblock \emph{arXiv preprint arXiv:2410.24164}, 2024.

\bibitem[Black et~al.(2025)Black, Brown, Darpinian, Dhabalia, Driess, Esmail,
  Equi, Finn, Fusai, Galliker, et~al.]{black2025pi_}
K.~Black, N.~Brown, J.~Darpinian, K.~Dhabalia, D.~Driess, A.~Esmail, M.~R.
  Equi, C.~Finn, N.~Fusai, M.~Y. Galliker, et~al.
\newblock $\pi_{0.5}$: a vision-language-action model with open-world
  generalization.
\newblock In \emph{CoRL}, 2025.

\bibitem[Chi et~al.(2025)Chi, Xu, Feng, Cousineau, Du, Burchfiel, Tedrake, and
  Song]{chi2025diffusion}
C.~Chi, Z.~Xu, S.~Feng, E.~Cousineau, Y.~Du, B.~Burchfiel, R.~Tedrake, and
  S.~Song.
\newblock Diffusion policy: Visuomotor policy learning via action diffusion.
\newblock \emph{IJRR}, 2025.

\bibitem[Kim et~al.(2025)Kim, Pertsch, Karamcheti, Xiao, Balakrishna, Nair,
  Rafailov, Foster, Sanketi, Vuong, et~al.]{kim2025openvla}
M.~J. Kim, K.~Pertsch, S.~Karamcheti, T.~Xiao, A.~Balakrishna, S.~Nair,
  R.~Rafailov, E.~P. Foster, P.~R. Sanketi, Q.~Vuong, et~al.
\newblock Openvla: An open-source vision-language-action model.
\newblock In \emph{CoRL}, 2025.

\bibitem[Ze et~al.(2024)Ze, Zhang, Zhang, Hu, Wang, and Xu]{ze20243d}
Y.~Ze, G.~Zhang, K.~Zhang, C.~Hu, M.~Wang, and H.~Xu.
\newblock 3d diffusion policy: Generalizable visuomotor policy learning via
  simple 3d representations.
\newblock In \emph{RSS}, 2024.

\bibitem[Zhao et~al.(2023)Zhao, Kumar, Levine, and Finn]{zhao2023learning}
T.~Zhao, V.~Kumar, S.~Levine, and C.~Finn.
\newblock Learning fine-grained bimanual manipulation with low-cost hardware.
\newblock In \emph{RSS}, 2023.

\bibitem[Brohan et~al.(2022)Brohan, Brown, Carbajal, Chebotar, Dabis, Finn,
  Gopalakrishnan, Hausman, Herzog, Hsu, Ibarz, Ichter, Irpan, Jackson,
  Jesmonth, Joshi, Julian, Kalashnikov, Kuang, Leal, Lee, Levine, Lu, Malla,
  Manjunath, Mordatch, Nachum, Parada, Peralta, Perez, Pertsch, Quiambao, Rao,
  Ryoo, Salazar, Sanketi, Sayed, Singh, Sontakke, Stone, Tan, Tran, Vanhoucke,
  Vega, Vuong, Xia, Xiao, Xu, Xu, Yu, and Zitkovich]{rt12022arxiv}
A.~Brohan, N.~Brown, J.~Carbajal, Y.~Chebotar, J.~Dabis, C.~Finn,
  K.~Gopalakrishnan, K.~Hausman, A.~Herzog, J.~Hsu, J.~Ibarz, B.~Ichter,
  A.~Irpan, T.~Jackson, S.~Jesmonth, N.~Joshi, R.~Julian, D.~Kalashnikov,
  Y.~Kuang, I.~Leal, K.-H. Lee, S.~Levine, Y.~Lu, U.~Malla, D.~Manjunath,
  I.~Mordatch, O.~Nachum, C.~Parada, J.~Peralta, E.~Perez, K.~Pertsch,
  J.~Quiambao, K.~Rao, M.~Ryoo, G.~Salazar, P.~Sanketi, K.~Sayed, J.~Singh,
  S.~Sontakke, A.~Stone, C.~Tan, H.~Tran, V.~Vanhoucke, S.~Vega, Q.~Vuong,
  F.~Xia, T.~Xiao, P.~Xu, S.~Xu, T.~Yu, and B.~Zitkovich.
\newblock Rt-1: Robotics transformer for real-world control at scale.
\newblock In \emph{arXiv preprint arXiv:2212.06817}, 2022.

\bibitem[Brohan et~al.(2023)Brohan, Brown, Carbajal, Chebotar, Chen,
  Choromanski, Ding, Driess, Dubey, Finn, Florence, Fu, Arenas, Gopalakrishnan,
  Han, Hausman, Herzog, Hsu, Ichter, Irpan, Joshi, Julian, Kalashnikov, Kuang,
  Leal, Lee, Lee, Levine, Lu, Michalewski, Mordatch, Pertsch, Rao, Reymann,
  Ryoo, Salazar, Sanketi, Sermanet, Singh, Singh, Soricut, Tran, Vanhoucke,
  Vuong, Wahid, Welker, Wohlhart, Wu, Xia, Xiao, Xu, Xu, Yu, and
  Zitkovich]{rt22023arxiv}
A.~Brohan, N.~Brown, J.~Carbajal, Y.~Chebotar, X.~Chen, K.~Choromanski,
  T.~Ding, D.~Driess, A.~Dubey, C.~Finn, P.~Florence, C.~Fu, M.~G. Arenas,
  K.~Gopalakrishnan, K.~Han, K.~Hausman, A.~Herzog, J.~Hsu, B.~Ichter,
  A.~Irpan, N.~Joshi, R.~Julian, D.~Kalashnikov, Y.~Kuang, I.~Leal, L.~Lee,
  T.-W.~E. Lee, S.~Levine, Y.~Lu, H.~Michalewski, I.~Mordatch, K.~Pertsch,
  K.~Rao, K.~Reymann, M.~Ryoo, G.~Salazar, P.~Sanketi, P.~Sermanet, J.~Singh,
  A.~Singh, R.~Soricut, H.~Tran, V.~Vanhoucke, Q.~Vuong, A.~Wahid, S.~Welker,
  P.~Wohlhart, J.~Wu, F.~Xia, T.~Xiao, P.~Xu, S.~Xu, T.~Yu, and B.~Zitkovich.
\newblock Rt-2: Vision-language-action models transfer web knowledge to robotic
  control.
\newblock In \emph{arXiv preprint arXiv:2307.15818}, 2023.

\bibitem[{Octo Model Team} et~al.(2024){Octo Model Team}, Ghosh, Walke,
  Pertsch, Black, Mees, Dasari, Hejna, Xu, Luo, Kreiman, Tan, Chen, Sanketi,
  Vuong, Xiao, Sadigh, Finn, and Levine]{octo_2023}
{Octo Model Team}, D.~Ghosh, H.~Walke, K.~Pertsch, K.~Black, O.~Mees,
  S.~Dasari, J.~Hejna, C.~Xu, J.~Luo, T.~Kreiman, Y.~Tan, L.~Y. Chen,
  P.~Sanketi, Q.~Vuong, T.~Xiao, D.~Sadigh, C.~Finn, and S.~Levine.
\newblock Octo: An open-source generalist robot policy.
\newblock In \emph{Proceedings of Robotics: Science and Systems}, Delft,
  Netherlands, 2024.

\bibitem[Liu et~al.(2024)Liu, Wu, Li, Tan, Chen, Wang, Xu, Su, and
  Zhu]{liu2024rdt}
S.~Liu, L.~Wu, B.~Li, H.~Tan, H.~Chen, Z.~Wang, K.~Xu, H.~Su, and J.~Zhu.
\newblock Rdt-1b: a diffusion foundation model for bimanual manipulation.
\newblock \emph{arXiv preprint arXiv:2410.07864}, 2024.

\bibitem[Kim et~al.(2024)Kim, Pertsch, Karamcheti, Xiao, Balakrishna, Nair,
  Rafailov, Foster, Lam, Sanketi, Vuong, Kollar, Burchfiel, Tedrake, Sadigh,
  Levine, Liang, and Finn]{kim24openvla}
M.~Kim, K.~Pertsch, S.~Karamcheti, T.~Xiao, A.~Balakrishna, S.~Nair,
  R.~Rafailov, E.~Foster, G.~Lam, P.~Sanketi, Q.~Vuong, T.~Kollar,
  B.~Burchfiel, R.~Tedrake, D.~Sadigh, S.~Levine, P.~Liang, and C.~Finn.
\newblock Openvla: An open-source vision-language-action model.
\newblock \emph{arXiv preprint arXiv:2406.09246}, 2024.

\bibitem[Yang et~al.(2025)Yang, Yu, Wu, Yan, Li, Cheng, Zou, Fang, Yin, Liu,
  Han, Lu, and Wang]{yang2025egovlalearningvisionlanguageactionmodels}
R.~Yang, Q.~Yu, Y.~Wu, R.~Yan, B.~Li, A.-C. Cheng, X.~Zou, Y.~Fang, H.~Yin,
  S.~Liu, S.~Han, Y.~Lu, and X.~Wang.
\newblock Egovla: Learning vision-language-action models from egocentric human
  videos, 2025.
\newblock URL \url{https://arxiv.org/abs/2507.12440}.

\bibitem[Bi et~al.(2025)Bi, Tan, Xie, Wang, Huang, Liu, Zhao, Feng, Xiang,
  Rong, et~al.]{bi2025motus}
H.~Bi, H.~Tan, S.~Xie, Z.~Wang, S.~Huang, H.~Liu, R.~Zhao, Y.~Feng, C.~Xiang,
  Y.~Rong, et~al.
\newblock Motus: A unified latent action world model.
\newblock \emph{arXiv preprint arXiv:2512.13030}, 2025.

\bibitem[Kim et~al.(2026)Kim, Gao, Lin, Lin, Ge, Lam, Liang, Song, Liu, Finn,
  and Gu]{kim2026cosmos}
M.~J. Kim, Y.~Gao, T.-Y. Lin, Y.-C. Lin, Y.~Ge, G.~Lam, P.~Liang, S.~Song,
  M.-Y. Liu, C.~Finn, and J.~Gu.
\newblock Cosmos policy: Fine-tuning video models for visuomotor control and
  planning.
\newblock \emph{arXiv preprint arXiv:2601.16163}, 2026.

\bibitem[Ye et~al.(2026)Ye, Ge, Zheng, Gao, Yu, Kurian, Indupuru, Tan, Zhu,
  Xiang, et~al.]{ye2026world}
S.~Ye, Y.~Ge, K.~Zheng, S.~Gao, S.~Yu, G.~Kurian, S.~Indupuru, Y.~L. Tan,
  C.~Zhu, J.~Xiang, et~al.
\newblock World action models are zero-shot policies.
\newblock \emph{arXiv preprint arXiv:2602.15922}, 2026.

\bibitem[Yuan et~al.(2026)Yuan, Dong, Liu, and Zhao]{yuan2026fastwam}
T.~Yuan, Z.~Dong, Y.~Liu, and H.~Zhao.
\newblock Fast-wam: Do world action models need test-time future imagination?
\newblock \emph{arXiv preprint arXiv:2603.16666}, 2026.

\bibitem[Li et~al.(2026)Li, Zhang, Luo, Yang, Wang, Han, Yu, Gao, Xue, Zhu,
  Shen, and Xu]{lingbot-va2026}
L.~Li, Q.~Zhang, Y.~Luo, S.~Yang, R.~Wang, F.~Han, M.~Yu, Z.~Gao, N.~Xue,
  X.~Zhu, Y.~Shen, and Y.~Xu.
\newblock Causal world modeling for robot control.
\newblock \emph{arXiv preprint arXiv:2601.21998}, 2026.

\bibitem[Bharadhwaj et~al.(2024)Bharadhwaj, Dwibedi, Gupta, Tulsiani, Doersch,
  Xiao, Shah, Xia, Sadigh, and Kirmani]{bharadhwaj2024gen2act}
H.~Bharadhwaj, D.~Dwibedi, A.~Gupta, S.~Tulsiani, C.~Doersch, T.~Xiao, D.~Shah,
  F.~Xia, D.~Sadigh, and S.~Kirmani.
\newblock Gen2act: Human video generation in novel scenarios enables
  generalizable robot manipulation.
\newblock \emph{arXiv preprint arXiv:2409.16283}, 2024.

\bibitem[Liang et~al.(2025)Liang, Tokmakov, Liu, Sudhakar, Shah, Ambrus, and
  Vondrick]{liang2025video}
J.~Liang, P.~Tokmakov, R.~Liu, S.~Sudhakar, P.~Shah, R.~Ambrus, and
  C.~Vondrick.
\newblock Video generators are robot policies.
\newblock \emph{arXiv preprint arXiv:2508.00795}, 2025.

\bibitem[Li et~al.(2025)Li, Gao, Sadigh, and Song]{li2025unified}
S.~Li, Y.~Gao, D.~Sadigh, and S.~Song.
\newblock Unified video action model.
\newblock \emph{arXiv preprint arXiv:2503.00200}, 2025.

\bibitem[Liao et~al.(2025)Liao, Zhou, Huang, Yang, Chen, Jiang, Hu, Cai, Liu,
  Luo, Chen, Yan, Yao, and Ren]{liao2025genie}
Y.~Liao, P.~Zhou, S.~Huang, D.~Yang, S.~Chen, Y.~Jiang, Y.~Hu, J.~Cai, S.~Liu,
  J.~Luo, L.~Chen, S.~Yan, M.~Yao, and G.~Ren.
\newblock Genie envisioner: A unified world foundation platform for robotic
  manipulation.
\newblock \emph{arXiv preprint arXiv:2508.05635}, 2025.

\bibitem[Chen et~al.(2025)Chen, Zhang, Geng, Song, Freeman, Malik, Tedrake,
  Sitzmann, and Du]{chen2025largevideoplanner}
B.~Chen, T.~Zhang, H.~Geng, K.~Song, W.~T. Freeman, J.~Malik, R.~Tedrake,
  V.~Sitzmann, and Y.~Du.
\newblock Large video planner, 2025.
\newblock URL \url{http://arxiv.org/abs/2512.15840}.

\bibitem[Wang et~al.(2019)Wang, Kurutach, Liu, Abbeel, and Tamar]{Tamar-RSS-19}
A.~Wang, T.~Kurutach, K.~Liu, P.~Abbeel, and A.~Tamar.
\newblock Learning robotic manipulation through visual planning and acting.
\newblock In \emph{Proceedings of Robotics: Science and Systems},
  FreiburgimBreisgau, Germany, June 2019.
\newblock \doi{10.15607/RSS.2019.XV.074}.

\bibitem[Lou et~al.(2026)Lou, Chi, Zhang, Qian, Li, Zhang, Lyu, Song, Fu, Xu,
  Wang, and Zhang]{lou2026mask}
Y.~Lou, X.~Chi, X.~Zhang, Z.~Qian, C.~Li, R.~Zhang, Y.~Lyu, G.~Song, C.~Fu,
  H.~Xu, P.~Wang, and S.~Z. Zhang.
\newblock Mask world model: Predicting what matters for robust robot policy
  learning.
\newblock \emph{arXiv preprint arXiv:2604.19683v2}, 2026.

\bibitem[Assran et~al.(2025)Assran, Bardes, Fan, Garrido, Howes, Komeili,
  Muckley, Rizvi, Roberts, Sinha, Zholus, Arnaud, Gejji, Martin, Robert~Hogan,
  Dugas, Bojanowski, Khalidov, Labatut, Massa, Szafraniec, Krishnakumar, Li,
  Ma, Chandar, Meier, LeCun, Rabbat, and Ballas]{assran2025vjepa2}
M.~Assran, A.~Bardes, D.~Fan, Q.~Garrido, R.~Howes, M.~Komeili, M.~Muckley,
  A.~Rizvi, C.~Roberts, K.~Sinha, A.~Zholus, S.~Arnaud, A.~Gejji, A.~Martin,
  F.~Robert~Hogan, D.~Dugas, P.~Bojanowski, V.~Khalidov, P.~Labatut, F.~Massa,
  M.~Szafraniec, K.~Krishnakumar, Y.~Li, X.~Ma, S.~Chandar, F.~Meier, Y.~LeCun,
  M.~Rabbat, and N.~Ballas.
\newblock V-jepa~2: Self-supervised video models enable understanding,
  prediction and planning.
\newblock \emph{arXiv preprint arXiv:2506.09985}, 2025.

\bibitem[Terver et~al.(2025)Terver, Yang, Ponce, Bardes, and
  LeCun]{terver2025drivessuccessphysicalplanning}
B.~Terver, T.-Y. Yang, J.~Ponce, A.~Bardes, and Y.~LeCun.
\newblock What drives success in physical planning with joint-embedding
  predictive world models?, 2025.
\newblock URL \url{https://arxiv.org/abs/2512.24497}.

\bibitem[Maes et~al.(2026)Maes, Le~Lidec, Scieur, LeCun, and
  Balestriero]{leworldmodel2026}
L.~Maes, Q.~Le~Lidec, D.~Scieur, Y.~LeCun, and R.~Balestriero.
\newblock Leworldmodel: Stable end-to-end joint-embedding predictive
  architecture from pixels, 2026.

\bibitem[Chen et~al.(2025)Chen, Chen, Chen, Cai, Liu, Li, Liang, Lin, Ge, Gu,
  et~al.]{chen2025robotwin}
T.~Chen, Z.~Chen, B.~Chen, Z.~Cai, Y.~Liu, Z.~Li, Q.~Liang, X.~Lin, Y.~Ge,
  Z.~Gu, et~al.
\newblock Robotwin 2.0: A scalable data generator and benchmark with strong
  domain randomization for robust bimanual robotic manipulation.
\newblock \emph{arXiv preprint arXiv:2506.18088}, 2025.

\bibitem[Ha and Schmidhuber(2018)]{ha2018worldmodels}
D.~Ha and J.~Schmidhuber.
\newblock Recurrent world models facilitate policy evolution.
\newblock In \emph{Advances in Neural Information Processing Systems 31}, pages
  2451--2463. Curran Associates, Inc., 2018.
\newblock URL
  \url{https://papers.nips.cc/paper/7512-recurrent-world-models-facilitate-policy-evolution}.
\newblock \url{https://worldmodels.github.io}.

\bibitem[Hafner et~al.(2019)Hafner, Lillicrap, Ba, and
  Norouzi]{hafner2019dreamer}
D.~Hafner, T.~Lillicrap, J.~Ba, and M.~Norouzi.
\newblock Dream to control: Learning behaviors by latent imagination.
\newblock \emph{arXiv preprint arXiv:1912.01603}, 2019.

\bibitem[Hafner et~al.(2020)Hafner, Lillicrap, Norouzi, and
  Ba]{hafner2020dreamerv2}
D.~Hafner, T.~Lillicrap, M.~Norouzi, and J.~Ba.
\newblock Mastering atari with discrete world models.
\newblock \emph{arXiv preprint arXiv:2010.02193}, 2020.

\bibitem[Hafner et~al.(2025{\natexlab{a}})Hafner, Pasukonis, Ba, and
  Lillicrap]{hafner2025dreamerv3}
D.~Hafner, J.~Pasukonis, J.~Ba, and T.~Lillicrap.
\newblock Mastering diverse control tasks through world models.
\newblock \emph{Nature}, pages 1--7, 2025{\natexlab{a}}.

\bibitem[Hafner et~al.(2025{\natexlab{b}})Hafner, Yan, and
  Lillicrap]{hafner2025training}
D.~Hafner, W.~Yan, and T.~Lillicrap.
\newblock Training agents inside of scalable world models.
\newblock \emph{arXiv preprint arXiv:2509.24527}, 2025{\natexlab{b}}.
\newblock \doi{10.48550/arXiv.2509.24527}.
\newblock URL \url{https://arxiv.org/abs/2509.24527}.

\bibitem[Alonso et~al.(2024)Alonso, Jelley, Micheli, Kanervisto, Storkey,
  Pearce, and Fleuret]{alonso2024diffusionworldmodelingvisual}
E.~Alonso, A.~Jelley, V.~Micheli, A.~Kanervisto, A.~Storkey, T.~Pearce, and
  F.~Fleuret.
\newblock Diffusion for world modeling: Visual details matter in atari.
\newblock In \emph{Thirty-eighth Conference on Neural Information Processing
  Systems}, 2024.
\newblock URL \url{https://arxiv.org/abs/2405.12399}.

\bibitem[Lee et~al.(2026)Lee, Jung, Chun, Lee, Cai, Huang, Talreja, Dao, Liang,
  Huang, and Huang]{lee2026tracegen}
S.~Lee, Y.~Jung, I.~Chun, Y.-C. Lee, Z.~Cai, H.~Huang, A.~Talreja, T.~D. Dao,
  Y.~Liang, J.-B. Huang, and F.~Huang.
\newblock Tracegen: World modeling in 3d trace space enables learning from
  cross-embodiment videos.
\newblock In \emph{Proceedings of the IEEE/CVF Conference on Computer Vision
  and Pattern Recognition}, 2026.

\bibitem[Finn and Levine(2017)]{finn2017deep}
C.~Finn and S.~Levine.
\newblock Deep visual foresight for planning robot motion.
\newblock In \emph{2017 IEEE International Conference on Robotics and
  Automation (ICRA)}, 2017.

\bibitem[Ebert et~al.(2018)Ebert, Finn, Dasari, Xie, Lee, and
  Levine]{visualforesight}
F.~Ebert, C.~Finn, S.~Dasari, A.~Xie, A.~Lee, and S.~Levine.
\newblock Visual foresight: Model-based deep reinforcement learning for
  vision-based robotic control.
\newblock \emph{arXiv preprint arXiv:1812.00568}, 2018.

\bibitem[Yang et~al.(2023)Yang, Du, Ghasemipour, Tompson, Schuurmans, and
  Abbeel]{yang2023learning}
M.~Yang, Y.~Du, K.~Ghasemipour, J.~Tompson, D.~Schuurmans, and P.~Abbeel.
\newblock Learning interactive real-world simulators.
\newblock \emph{arXiv preprint arXiv:2310.06114}, 2023.

\bibitem[Zhou et~al.(2024)Zhou, Pan, LeCun, and
  Pinto]{zhou2024dinowmworldmodelspretrained}
G.~Zhou, H.~Pan, Y.~LeCun, and L.~Pinto.
\newblock Dino-wm: World models on pre-trained visual features enable zero-shot
  planning, 2024.
\newblock URL \url{https://arxiv.org/abs/2411.04983}.

\bibitem[NVIDIA et~al.(2025)NVIDIA, Agarwal, Ali, Bala, Balaji, Barker, Cai,
  Chattopadhyay, Chen, Cui, Ding, Dworakowski, Fan, Fenzi, Ferroni, Fidler,
  Fox, Ge, Ge, Gu, Gururani, He, Huang, Huffman, Jannaty, Jin, Kim, Kl\'{a}r,
  Lam, Lan, Leal-Taixe, Li, Li, Lin, Lin, Ling, Liu, Liu, Luo, Ma, Mao, Mo,
  Mousavian, Nah, Niverty, Page, Paschalidou, Patel, Pavao, Ramezanali, Reda,
  Ren, Sabavat, Schmerling, Shi, Stefaniak, Tang, Tchapmi, Tredak, Tseng,
  Varghese, Wang, Wang, Wang, Wang, Wei, Wei, Wu, Xu, Yang, Yen-Chen, Zeng,
  Zeng, Zhang, Zhang, Zhang, Zhao, and
  Zolkowski]{nvidia2025cosmosworldfoundationmodel}
NVIDIA, N.~Agarwal, A.~Ali, M.~Bala, Y.~Balaji, E.~Barker, T.~Cai,
  P.~Chattopadhyay, Y.~Chen, Y.~Cui, Y.~Ding, D.~Dworakowski, J.~Fan, M.~Fenzi,
  F.~Ferroni, S.~Fidler, D.~Fox, S.~Ge, Y.~Ge, J.~Gu, S.~Gururani, E.~He,
  J.~Huang, J.~Huffman, P.~Jannaty, J.~Jin, S.~W. Kim, G.~Kl\'{a}r, G.~Lam,
  S.~Lan, L.~Leal-Taixe, A.~Li, Z.~Li, C.-H. Lin, T.-Y. Lin, H.~Ling, M.-Y.
  Liu, X.~Liu, A.~Luo, Q.~Ma, H.~Mao, K.~Mo, A.~Mousavian, S.~Nah, S.~Niverty,
  D.~Page, D.~Paschalidou, Z.~Patel, L.~Pavao, M.~Ramezanali, F.~Reda, X.~Ren,
  V.~R.~N. Sabavat, E.~Schmerling, S.~Shi, B.~Stefaniak, S.~Tang, L.~Tchapmi,
  P.~Tredak, W.-C. Tseng, J.~Varghese, H.~Wang, H.~Wang, H.~Wang, T.-C. Wang,
  F.~Wei, X.~Wei, J.~Z. Wu, J.~Xu, W.~Yang, L.~Yen-Chen, X.~Zeng, Y.~Zeng,
  J.~Zhang, Q.~Zhang, Y.~Zhang, Q.~Zhao, and A.~Zolkowski.
\newblock Cosmos world foundation model platform for physical ai.
\newblock \emph{arXiv preprint arXiv:2501.03575}, 2025.

\bibitem[NVIDIA(2025)]{nvidia2025cosmospredict2}
NVIDIA.
\newblock Cosmos-predict2: World simulation model for physical ai, 2025.
\newblock URL \url{https://github.com/nvidia-cosmos/cosmos-predict2}.

\bibitem[Zeng et~al.(2019)Zeng, Luo, Suo, Sadat, Yang, Casas, and
  Urtasun]{Zeng_2019_CVPR}
W.~Zeng, W.~Luo, S.~Suo, A.~Sadat, B.~Yang, S.~Casas, and R.~Urtasun.
\newblock End-to-end interpretable neural motion planner.
\newblock In \emph{Proceedings of the IEEE/CVF Conference on Computer Vision
  and Pattern Recognition (CVPR)}, June 2019.

\bibitem[Zeng et~al.(2020)Zeng, Wang, Liao, Chen, Yang, and Urtasun]{Zeng2020}
W.~Zeng, S.~Wang, R.~Liao, Y.~Chen, B.~Yang, and R.~Urtasun.
\newblock Dsdnet: Deep structured self-driving network.
\newblock \emph{Proceedings of the European Conference on Computer Vision
  (ECCV)}, 2020.

\bibitem[Bansal et~al.(2019)Bansal, Krizhevsky, and Ogale]{Bansal_2019_RSS}
M.~Bansal, A.~Krizhevsky, and A.~Ogale.
\newblock Chauffeurnet: Learning to drive by imitating the best and
  synthesizing the worst.
\newblock In \emph{Proceedings of Robotics: Science and Systems (RSS)}, June
  2019.

\bibitem[Hu et~al.(2023)Hu, Yang, Chen, Li, Sima, Zhu, Chai, Du, Lin, Wang, Lu,
  Jia, Liu, Dai, Qiao, and Li]{hu2023_uniad}
Y.~Hu, J.~Yang, L.~Chen, K.~Li, C.~Sima, X.~Zhu, S.~Chai, S.~Du, T.~Lin,
  W.~Wang, L.~Lu, X.~Jia, Q.~Liu, J.~Dai, Y.~Qiao, and H.~Li.
\newblock Planning-oriented autonomous driving.
\newblock In \emph{Proceedings of the IEEE/CVF Conference on Computer Vision
  and Pattern Recognition}, 2023.

\bibitem[Finn et~al.(2016)Finn, Tan, Duan, Darrell, Levine, and
  Abbeel]{finn2016deep}
C.~Finn, X.~Y. Tan, Y.~Duan, T.~Darrell, S.~Levine, and P.~Abbeel.
\newblock Deep spatial autoencoders for visuomotor learning.
\newblock In \emph{2016 IEEE International Conference on Robotics and
  Automation (ICRA)}. IEEE, 2016.

\bibitem[Manuelli et~al.(2019)Manuelli, Gao, Florence, and
  Tedrake]{manuelli2022kpam}
L.~Manuelli, W.~Gao, P.~Florence, and R.~Tedrake.
\newblock {kPAM}: {KeyPoint} affordances for category-level robotic
  manipulation.
\newblock In \emph{Robotics Research}, 2019.
\newblock URL \url{https://arxiv.org/abs/1903.06684}.

\bibitem[Qin et~al.(2019)Qin, Fang, Zhu, Fei-Fei, and Savarese]{qin2019keto}
Z.~Qin, K.~Fang, Y.~Zhu, L.~Fei-Fei, and S.~Savarese.
\newblock Keto: Learning keypoint representations for tool manipulation, 2019.

\bibitem[Florence et~al.(2018)Florence, Manuelli, and
  Tedrake]{pmlr-v87-florence18a}
P.~R. Florence, L.~Manuelli, and R.~Tedrake.
\newblock Dense object nets: Learning dense visual object descriptors by and
  for robotic manipulation.
\newblock In A.~Billard, A.~Dragan, J.~Peters, and J.~Morimoto, editors,
  \emph{Proceedings of The 2nd Conference on Robot Learning}, volume~87 of
  \emph{Proceedings of Machine Learning Research}, pages 373--385. PMLR, 29--31
  Oct 2018.
\newblock URL \url{https://proceedings.mlr.press/v87/florence18a.html}.

\bibitem[Simeonov et~al.(2022)Simeonov, Du, Tagliasacchi, Tenenbaum, Rodriguez,
  Agrawal, and Sitzmann]{simeonovdu2021ndf}
A.~Simeonov, Y.~Du, A.~Tagliasacchi, J.~B. Tenenbaum, A.~Rodriguez, P.~Agrawal,
  and V.~Sitzmann.
\newblock Neural descriptor fields: Se(3)-equivariant object representations
  for manipulation.
\newblock In \emph{ICRA}, 2022.

\bibitem[Zeng et~al.(2020)Zeng, Florence, Tompson, Welker, Chien, Attarian,
  Armstrong, Krasin, Duong, Sindhwani, and Lee]{zeng2020transporter}
A.~Zeng, P.~Florence, J.~Tompson, S.~Welker, J.~Chien, M.~Attarian,
  T.~Armstrong, I.~Krasin, D.~Duong, V.~Sindhwani, and J.~Lee.
\newblock Transporter networks: Rearranging the visual world for robotic
  manipulation.
\newblock \emph{Conference on Robot Learning (CoRL)}, 2020.

\bibitem[Zhang and Ma(2025)]{Zhang_2025_BMVC}
J.~Zhang and K.~Ma.
\newblock Mask2act: Predictive multi-object tracking as video pre-training for
  robot manipulation.
\newblock In \emph{36th British Machine Vision Conference 2025, {BMVC} 2025,
  Sheffield, UK, November 24-27, 2025}. BMVA, 2025.
\newblock URL
  \url{https://bmva-archive.org.uk/bmvc/2025/assets/papers/Paper_124/paper.pdf}.

\bibitem[Ferraro et~al.(2025)Ferraro, Mazzaglia, Verbelen, and
  Dhoedt]{ferraro2025focus}
S.~Ferraro, P.~Mazzaglia, T.~Verbelen, and B.~Dhoedt.
\newblock {FOCUS}: Object-centric world models for robotic manipulation.
\newblock \emph{Frontiers in Neurorobotics}, 19:\penalty0 1585386, 2025.
\newblock \doi{10.3389/fnbot.2025.1585386}.

\bibitem[Zhu et~al.(2023)Zhu, Jiang, Stone, and Zhu]{zhu2023learning}
Y.~Zhu, Z.~Jiang, P.~Stone, and Y.~Zhu.
\newblock Learning generalizable manipulation policies with object-centric 3d
  representations.
\newblock In \emph{7th Annual Conference on Robot Learning}, 2023.

\bibitem[Shridhar et~al.(2022)Shridhar, Manuelli, and Fox]{shridhar2022peract}
M.~Shridhar, L.~Manuelli, and D.~Fox.
\newblock Perceiver-actor: A multi-task transformer for robotic manipulation.
\newblock In \emph{Proceedings of the 6th Conference on Robot Learning (CoRL)},
  2022.

\bibitem[Shridhar et~al.(2021)Shridhar, Manuelli, and Fox]{shridhar2021cliport}
M.~Shridhar, L.~Manuelli, and D.~Fox.
\newblock Cliport: What and where pathways for robotic manipulation.
\newblock In \emph{Proceedings of the 5th Conference on Robot Learning (CoRL)},
  2021.

\bibitem[Huang et~al.(2024)Huang, Wang, Li, Zhang, and Fei-Fei]{huang2024rekep}
W.~Huang, C.~Wang, Y.~Li, R.~Zhang, and L.~Fei-Fei.
\newblock Rekep: Spatio-temporal reasoning of relational keypoint constraints
  for robotic manipulation.
\newblock \emph{arXiv preprint arXiv:2409.01652}, 2024.

\bibitem[Mo et~al.(2021)Mo, Guibas, Mukadam, Gupta, and Tulsiani]{Mo_2021_ICCV}
K.~Mo, L.~J. Guibas, M.~Mukadam, A.~Gupta, and S.~Tulsiani.
\newblock Where2act: From pixels to actions for articulated 3d objects.
\newblock In \emph{Proceedings of the IEEE/CVF International Conference on
  Computer Vision (ICCV)}, pages 6813--6823, October 2021.

\bibitem[Tang et~al.(2025)Tang, Xiao, Deng, Hu, Dong, Zhang, Hsu, and
  Zhang]{tang2025mimicfunc}
C.~Tang, A.~Xiao, Y.~Deng, T.~Hu, W.~Dong, H.~Zhang, D.~Hsu, and H.~Zhang.
\newblock {MimicFunc}: Imitating tool manipulation from a single human video
  via functional correspondence.
\newblock In \emph{Proceedings of The 9th Conference on Robot Learning}, volume
  305, pages 4473--4492. PMLR, 2025.
\newblock URL \url{https://proceedings.mlr.press/v305/tang25a.html}.

\bibitem[Sim{\'e}oni et~al.(2025)Sim{\'e}oni, Vo, Seitzer, Baldassarre, Oquab,
  Jose, Khalidov, Szafraniec, Yi, Ramamonjisoa, et~al.]{simeoni2025dinov3}
O.~Sim{\'e}oni, H.~V. Vo, M.~Seitzer, F.~Baldassarre, M.~Oquab, C.~Jose,
  V.~Khalidov, M.~Szafraniec, S.~Yi, M.~Ramamonjisoa, et~al.
\newblock Dinov3.
\newblock \emph{arXiv preprint arXiv:2508.10104}, 2025.

\bibitem[Wu et~al.(2024)Wu, Shentu, Yi, Lin, and Abbeel]{wu2024gello}
P.~Wu, Y.~Shentu, Z.~Yi, X.~Lin, and P.~Abbeel.
\newblock Gello: A general, low-cost, and intuitive teleoperation framework for
  robot manipulators.
\newblock In \emph{IROS}, 2024.

\bibitem[Wan et~al.(2025)Wan, Wang, Ai, Wen, Mao, Xie, Chen, Yu, Zhao, Yang,
  Zeng, Wang, Zhang, Zhou, Wang, Chen, Zhu, Zhao, Yan, Huang, Feng, Zhang, Li,
  Wu, Chu, Feng, Zhang, Sun, Fang, Wang, Gui, Weng, Shen, Lin, Wang, Wang,
  Zhou, Wang, Shen, Yu, Shi, Huang, Xu, Kou, Lv, Li, Liu, Wang, Zhang, Huang,
  Li, Wu, Liu, Pan, Zheng, Hong, Shi, Feng, Jiang, Han, Wu, and Liu]{wan2025}
T.~Wan, A.~Wang, B.~Ai, B.~Wen, C.~Mao, C.-W. Xie, D.~Chen, F.~Yu, H.~Zhao,
  J.~Yang, J.~Zeng, J.~Wang, J.~Zhang, J.~Zhou, J.~Wang, J.~Chen, K.~Zhu,
  K.~Zhao, K.~Yan, L.~Huang, M.~Feng, N.~Zhang, P.~Li, P.~Wu, R.~Chu, R.~Feng,
  S.~Zhang, S.~Sun, T.~Fang, T.~Wang, T.~Gui, T.~Weng, T.~Shen, W.~Lin,
  W.~Wang, W.~Wang, W.~Zhou, W.~Wang, W.~Shen, W.~Yu, X.~Shi, X.~Huang, X.~Xu,
  Y.~Kou, Y.~Lv, Y.~Li, Y.~Liu, Y.~Wang, Y.~Zhang, Y.~Huang, Y.~Li, Y.~Wu,
  Y.~Liu, Y.~Pan, Y.~Zheng, Y.~Hong, Y.~Shi, Y.~Feng, Z.~Jiang, Z.~Han, Z.-F.
  Wu, and Z.~Liu.
\newblock Wan: Open and advanced large-scale video generative models.
\newblock \emph{arXiv preprint arXiv:2503.20314}, 2025.

\bibitem[Zakka(2026)]{zakka2026mink}
K.~Zakka.
\newblock {Mink}: Python inverse kinematics based on {MuJoCo}.
\newblock Software, version 1.1.0, 2026.
\newblock URL \url{https://github.com/kevinzakka/mink}.

\bibitem[Yuan et~al.(2023)Yuan, Murali, Mousavian, and Fox]{yuan2023m2t2}
W.~Yuan, A.~Murali, A.~Mousavian, and D.~Fox.
\newblock {M2T2}: Multi-task masked transformer for object-centric pick and
  place.
\newblock In \emph{Proceedings of The 7th Conference on Robot Learning}, volume
  229, pages 3619--3630. PMLR, 2023.
\newblock URL \url{https://proceedings.mlr.press/v229/yuan23a.html}.

\end{thebibliography}
\clearpage
\appendix
\section{Real-World OBB Ground-Truth Construction}
\label{sec:supp_real_obb_gt}

\method{} is supervised with the image-plane oriented bounding box (OBB) trajectory of the actively manipulated object. In simulation, this supervision is obtained directly from simulator object states. In the real world, we reconstruct OBB trajectories from recorded RGB-D observations, camera calibration, object meshes, and robot proprioception.

\paragraph{Pose Reconstruction}
For each demonstration, we first estimate the object pose while the object is visible before the grasp. We use SAM 3 to obtain a text-prompted object mask and FoundationPose to estimate the object's 6-DoF pose $T_{\mathrm{cam}\leftarrow\mathrm{obj}}$ from the corresponding object mesh. After grasping, the object is often occluded by the robot gripper, so direct vision-based pose estimation becomes unreliable. We therefore propagate the object pose using the robot end-effector trajectory.

Specifically, we detect the grasp frame $g$ from the gripper signal and choose a switch frame $s$ shortly after grasping. At this frame, we compute a fixed object-in-gripper transform:
\begin{equation}
    T_{\mathrm{ee}\leftarrow\mathrm{obj}}
    =
    T_{\mathrm{base}\leftarrow\mathrm{ee}}^{-1}
    T_{\mathrm{base}\leftarrow\mathrm{cam}}
    T_{\mathrm{cam}\leftarrow\mathrm{obj}}.
\end{equation}
During the grasped phase, we recover the object pose from end-effector kinematics:
\begin{equation}
    T_{\mathrm{cam}\leftarrow\mathrm{obj}}
    =
    T_{\mathrm{cam}\leftarrow\mathrm{base}}
    T_{\mathrm{base}\leftarrow\mathrm{ee}}
    T_{\mathrm{ee}\leftarrow\mathrm{obj}},
\end{equation}
where $T_{\mathrm{base}\leftarrow\mathrm{ee}}$ is the recorded end-effector pose.
After release, we keep the object pose fixed at the final propagated pose. This annotation pipeline assumes known meshes, calibrated RGB-D observations, and an approximately fixed object-to-gripper transform. Grasp slip, release motion, and errors in the initial pose can introduce systematic target errors. These assumptions apply to supervision construction; the deployed policy uses RGB images and proprioception. We do not report a quantitative reconstruction-error evaluation.

\paragraph{OBB Projection} Given the fused object pose, object mesh extents, and camera intrinsics, we project the eight 3D OBB corners into the camera image. The resulting 16-dimensional vector of pixel coordinates,
\[ x = [u_1,v_1,\ldots,u_8,v_8] \in \mathbb{R}^{16}, \] is used as the per-frame box state for real-world \method{} training.

\section{Real-World Progress Metrics}
\label{sec:supp_real_progress}

We evaluate real-world policies on three manipulation tasks: Make Coffee, Water Plant, and Place Pokémon. We report final task success rate and progress rate over ordered task-specific subgoals. For each rollout, progress is the fraction of the task's ordered subgoals completed; we average this fraction over rollouts and express it as a percentage.

\begin{table}[h]
    \centering
    \small
    \caption{Real-world progress subgoals. Progress is measured by the number of completed ordered subgoals in each rollout.}
    \label{tab:real_subgoals}
    \begin{tabular}{p{0.18\linewidth} p{0.12\linewidth} p{0.60\linewidth}}
        \toprule
        Task & \# Subgoals & Ordered subgoals \\
        \midrule
        Make Coffee & 5 &
        grasp mug; place mug; grasp K-Cup; place K-Cup; close lid \\
        Water Plant & 3 &
        grasp pitcher; water plant; place pitcher \\
        Place Pokemon & 2 &
        grasp toy; place toy \\
        \bottomrule
    \end{tabular}
\end{table}

\section{Additional Multi-Task Simulation Results}
\label{sec:supp_multitask}

We report the full per-task breakdown for the multi-task simulation setting in Table~\ref{tab:generalization}. Models are trained on 50 RoboTwin2.0 tasks with 50 demonstrations per task collected under clean setting, and evaluated for 100 trials with seed 0 under both clean and randomized settings.

\begin{table}[h]
    \centering
    \small
    \caption{Multi-task simulation summary. We report average success rate (\%) over RoboTwin2.0 tasks.}
    \label{tab:multitask_summary}
    \begin{tabular}{lccc}
        \toprule
        Method & Cameras & Clean Avg. & Randomized Avg. \\
        \midrule
        $\pi_{0.5}$ & 3 RGB & 52.6 & 26.9 \\
        $\pi_{0.5}$ + BoxWAM & 3 RGB & \textbf{61.5} & \textbf{34.0} \\
        \bottomrule
    \end{tabular}
\end{table}

\small
\begin{longtable}{lcc}
    \caption{Per-task multi-task simulation results under the clean setting trained with 2500 demos and joint based control. We report success rate (\%).}
    \label{tab:multitask_clean}
    \\
    \toprule
    Task & $\pi_{0.5}$ & $\pi_{0.5}$ + BoxWAM \\
    \midrule
    \endfirsthead

    \toprule
    Task & $\pi_{0.5}$ & $\pi_{0.5}$ + BoxWAM \\
    \midrule
    \endhead

    \bottomrule
    \endlastfoot

    adjust\_bottle & 68.0 & 94.0 \\
    beat\_block\_hammer & 81.0 & 86.0 \\
    blocks\_ranking\_rgb & 6.0 & 19.0 \\
    blocks\_ranking\_size & 28.0 & 46.0 \\
    click\_alarmclock & 89.0 & 93.0 \\
    click\_bell & 78.0 & 73.0 \\
    dump\_bin\_bigbin & 30.0 & 64.0 \\
    grab\_roller & 84.0 & 87.0 \\
    handover\_block & 14.0 & 15.0 \\
    handover\_mic & 61.0 & 22.0 \\
    hanging\_mug & 13.0 & 13.0 \\
    lift\_pot & 32.0 & 64.0 \\
    move\_can\_pot & 30.0 & 39.0 \\
    move\_pillbottle\_pad & 16.0 & 36.0 \\
    move\_playingcard\_away & 82.0 & 67.0 \\
    move\_stapler\_pad & 33.0 & 26.0 \\
    open\_laptop & 87.0 & 90.0 \\
    open\_microwave & 33.0 & 80.0 \\
    pick\_diverse\_bottles & 50.0 & 68.0 \\
    pick\_dual\_bottles & 66.0 & 88.0 \\
    place\_a2b\_left & 60.0 & 67.0 \\
    place\_a2b\_right & 55.0 & 55.0 \\
    place\_bread\_basket & 67.0 & 84.0 \\
    place\_bread\_skillet & 63.0 & 77.0 \\
    place\_burger\_fries & 65.0 & 94.0 \\
    place\_can\_basket & 27.0 & 49.0 \\
    place\_cans\_plasticbox & 60.0 & 92.0 \\
    place\_container\_plate & 83.0 & 95.0 \\
    place\_dual\_shoes & 2.0 & 4.0 \\
    place\_empty\_cup & 70.0 & 90.0 \\
    place\_fan & 36.0 & 36.0 \\
    place\_mouse\_pad & 37.0 & 49.0 \\
    place\_object\_basket & 66.0 & 67.0 \\
    place\_object\_scale & 30.0 & 48.0 \\
    place\_object\_stand & 81.0 & 83.0 \\
    place\_phone\_stand & 37.0 & 54.0 \\
    place\_shoe & 90.0 & 93.0 \\
    press\_stapler & 89.0 & 91.0 \\
    put\_bottles\_dustbin & 58.0 & 71.0 \\
    put\_object\_cabinet & 50.0 & 65.0 \\
    rotate\_qrcode & 39.0 & 66.0 \\
    scan\_object & 42.0 & 74.0 \\
    shake\_bottle\_horizontally & 99.0 & 100.0 \\
    shake\_bottle & 100.0 & 100.0 \\
    stack\_blocks\_three & 0.0 & 0.0 \\
    stack\_blocks\_two & 4.0 & 7.0 \\
    stack\_bowls\_three & 72.0 & 60.0 \\
    stack\_bowls\_two & 85.0 & 84.0 \\
    stamp\_seal & 23.0 & 18.0 \\
    turn\_switch & 57.0 & 30.0 \\
\end{longtable}

\begin{longtable}{lcc}
    \caption{Per-task multi-task simulation results under the randomized setting trained with 2500 demos and joint based control. We report success rate (\%).}
    \label{tab:multitask_randomized}
    \\
    \toprule
    Task & $\pi_{0.5}$ & $\pi_{0.5}$ + BoxWAM \\
    \midrule
    \endfirsthead

    \toprule
    Task & $\pi_{0.5}$ & $\pi_{0.5}$ + BoxWAM \\
    \midrule
    \endhead

    \bottomrule
    \endlastfoot

    adjust\_bottle & 15.0 & 53.0 \\
    beat\_block\_hammer & 20.0 & 9.0 \\
    blocks\_ranking\_rgb & 6.0 & 9.0 \\
    blocks\_ranking\_size & 10.0 & 19.0 \\
    click\_alarmclock & 48.0 & 62.0 \\
    click\_bell & 51.0 & 50.0 \\
    dump\_bin\_bigbin & 24.0 & 39.0 \\
    grab\_roller & 54.0 & 71.0 \\
    handover\_block & 1.0 & 0.0 \\
    handover\_mic & 3.0 & 0.0 \\
    hanging\_mug & 1.0 & 0.0 \\
    lift\_pot & 1.0 & 5.0 \\
    move\_can\_pot & 1.0 & 4.0 \\
    move\_pillbottle\_pad & 3.0 & 21.0 \\
    move\_playingcard\_away & 12.0 & 18.0 \\
    move\_stapler\_pad & 3.0 & 13.0 \\
    open\_laptop & 32.0 & 27.0 \\
    open\_microwave & 21.0 & 11.0 \\
    pick\_diverse\_bottles & 26.0 & 27.0 \\
    pick\_dual\_bottles & 42.0 & 40.0 \\
    place\_a2b\_left & 31.0 & 36.0 \\
    place\_a2b\_right & 24.0 & 36.0 \\
    place\_bread\_basket & 42.0 & 61.0 \\
    place\_bread\_skillet & 23.0 & 32.0 \\
    place\_burger\_fries & 58.0 & 78.0 \\
    place\_can\_basket & 8.0 & 25.0 \\
    place\_cans\_plasticbox & 25.0 & 71.0 \\
    place\_container\_plate & 76.0 & 65.0 \\
    place\_dual\_shoes & 3.0 & 1.0 \\
    place\_empty\_cup & 50.0 & 47.0 \\
    place\_fan & 17.0 & 22.0 \\
    place\_mouse\_pad & 8.0 & 8.0 \\
    place\_object\_basket & 28.0 & 36.0 \\
    place\_object\_scale & 17.0 & 26.0 \\
    place\_object\_stand & 51.0 & 43.0 \\
    place\_phone\_stand & 4.0 & 11.0 \\
    place\_shoe & 70.0 & 61.0 \\
    press\_stapler & 64.0 & 73.0 \\
    put\_bottles\_dustbin & 29.0 & 48.0 \\
    put\_object\_cabinet & 27.0 & 15.0 \\
    rotate\_qrcode & 7.0 & 11.0 \\
    scan\_object & 22.0 & 61.0 \\
    shake\_bottle\_horizontally & 93.0 & 96.0 \\
    shake\_bottle & 90.0 & 97.0 \\
    stack\_blocks\_three & 0.0 & 27.0 \\
    stack\_blocks\_two & 1.0 & 41.0 \\
    stack\_bowls\_three & 33.0 & 35.0 \\
    stack\_bowls\_two & 52.0 & 41.0 \\
    stamp\_seal & 10.0 & 4.0 \\
    turn\_switch & 7.0 & 13.0 \\
\end{longtable}
\normalsize

\end{document}